%% file: arxiv.tex
\documentclass[final,5p,times,twocolumn]{elsarticle}

\usepackage{booktabs} % for toprule, midrule, bottomrule in tables
\usepackage{graphicx} % to inlcude graphics with \includegraphics
\usepackage[table]{xcolor}
\usepackage{amsmath,amssymb,mathrsfs}
\usepackage{algorithm,algpseudocode,color,float,listings,pifont,makecell,multirow,soul,subcaption,tcolorbox,wrapfig}
\usepackage{tabularx}
    \newcolumntype{C}{>{\centering\arraybackslash}X}
    \newcolumntype{R}{>{\raggedleft\arraybackslash}X}
\usepackage{hyperref}
\usepackage{ifthen}
    \newboolean{arxiv}\setboolean{arxiv}{true}

\newtcolorbox{mybox}[1]{colback=red!5!white,colframe=red!75!black,fonttitle=\bfseries,title=#1}

\newcommand{\cB}{\mathcal{B}}
\newcommand{\cD}{\mathcal{D}}
\newcommand{\bbE}{\mathbb{E}}

\newcommand{\bbR}{\mathbb{R}}
\newcommand{\bbI}{\mathbb{I}}

\makeatletter
\newcommand{\onedotA}{\ifx\@let@token.\else.\null\fi\xspace}
\DeclareRobustCommand\onedot{\futurelet\@let@token\onedotA}

\makeatother

\DeclareMathOperator*{\argmin}{arg\,min}

\usepackage[upgrade=true]{acro}
\DeclareAcronym{BN}{short=BN, long=batch normalisation}
\DeclareAcronym{CORAL}{short=CORAL, long=CORrelation ALignment}
\DeclareAcronym{DANN}{short=DANN, long=domain-adversarial neural network}
\DeclareAcronym{DNN}{short=DNN, long=deep neural network}
\DeclareAcronym{DUA}{short=DUA, long=Dynamic Unsupervised Adaptation}
\DeclareAcronym{FP}{short=FP, long=false positive}
\DeclareAcronym{MMD}{short=MMD, long=maximum mean discrepancy}
\DeclareAcronym{SDA}{short=SDA, long=supervised domain adaptation, first-long-format=\emph}
\DeclareAcronym{SFDA}{short=SFDA, long=source-free domain adaptation, first-long-format=\emph}
\DeclareAcronym{SSDA}{short=SSDA, long=semi-supervised domain adaptation}
\DeclareAcronym{TTA}{short=TTA, long=test-time adaptation, first-long-format=\emph}
\DeclareAcronym{UDA}{short=UDA, long=unsupervised domain adaptation, first-long-format=\emph}
\begin{document}

\begin{frontmatter}

\title{Domain Adaptation for Satellite-Borne Hyperspectral Cloud Detection}

\author[1]{Andrew Du\corref{cor1}}
    \cortext[cor1]{Corresponding author}
    \ead{andrew.du@adelaide.edu.au}
\author[1]{Anh-Dzung Doan}
\author[2]{Yee Wei Law}
\author[1]{Tat-Jun Chin}

\affiliation[1]{organization={The University of Adelaide},
            % addressline={Address One}, 
            city={Adelaide},
            postcode={5000}, 
            state={South Australia},
            country={Australia}}
\affiliation[2]{organization={University of South Australia},
            % addressline={Address One}, 
            city={Mawson Lakes},
            postcode={5095}, 
            state={South Australia},
            country={Australia}}

\input{abstract}


\end{frontmatter}

\input{main}

\bibliographystyle{elsarticle-num} 
\bibliography{IEEEabrv,references}

\end{document}

%% file: abstract.tex
\begin{abstract}
%Hyperspectral imaging supplies valuable data at high spatio-temporal-spectral resolutions, but the resultant large data volumes readily outstrip the storage and communication capacities of a typical Earth observation satellite. For conserving storage and bandwidth, a satellite can estimate cloud coverage in images based on an onboard cloud detection model, and downlink only images with low cloud coverage. The advent of satellite-borne hardware accelerators enables the usage of deep architectures and features, such as those of convolution neural networks (CNNs), for accurate cloud detection. However, new missions that employ new sensors typically do not have enough representative datasets to train a CNN model, and a model trained solely on data from the previous missions tend to underperform in the new missions. This performance gap stems from the domain gap, i.e., difference in how data is distributed in the training/source and test/target domains. To bridge the domain gap, this article contributes two methods: (i) a bandwidth-efficient supervised domain adaptation (SDA) method that combines refinement with sparse updating, and (ii) a test-time adaptation (TTA) method. The proposed SDA method (i) downlinks a small amount of new data, (ii) refines an offline ground-based CNN model based on the new data, and (iii) uplinks the refined weights to update the online space-based model. Compared to a naive SDA method, the proposed SDA method can save \hl{XXX} MB of overhead per model update. The proposed TTA method ....
The advent of satellite-borne machine learning hardware accelerators has enabled the on-board processing of payload data using machine learning techniques such as convolutional neural networks (CNN). A notable example is using a CNN to detect the presence of clouds in hyperspectral data captured on Earth observation (EO) missions, whereby only clear sky data is downlinked to conserve bandwidth. However, prior to deployment, new missions that employ new sensors will not have enough representative datasets to train a CNN model, while a model trained solely on data from previous missions will underperform when deployed to process the data on the new missions. This underperformance stems from the domain gap, i.e., differences in the underlying distributions of the data generated by the different sensors in previous and future missions. In this paper, we address the domain gap problem in the context of on-board hyperspectral cloud detection. Our main contributions lie in formulating new domain adaptation tasks that are motivated by a concrete EO mission, developing a novel algorithm for bandwidth-efficient supervised domain adaptation, and demonstrating test-time adaptation algorithms on space deployable neural network accelerators. Our contributions enable minimal data transmission to be invoked (e.g., only 1\% of the weights in ResNet50) to achieve domain adaptation, thereby allowing more sophisticated CNN models to be deployed and updated on satellites without being hampered by domain gap and bandwidth limitations.
\end{abstract}

%% file: main.tex
\section{Introduction}\label{sec:introduction}

% EO satellites and their usefulness
Space provides a useful vantage point for monitoring large-scale trends on the surface of the Earth~\cite{manfreda2018use}. For that reason, numerous Earth observation (EO) satellite missions have been launched or are being planned. Typical EO satellites carry multispectral or hyperspectral sensors that measure the electromagnetic radiations emitted or reflected from the surface, which are then processed to form \emph{data cubes}. These data cubes are the valuable inputs to various EO applications.

\begin{figure}[ht]\centering
\ifarxiv
    \begin{subfigure}{\columnwidth}
\else
    \begin{subfigure}{0.8\columnwidth}
\fi
\includegraphics[width=\linewidth]{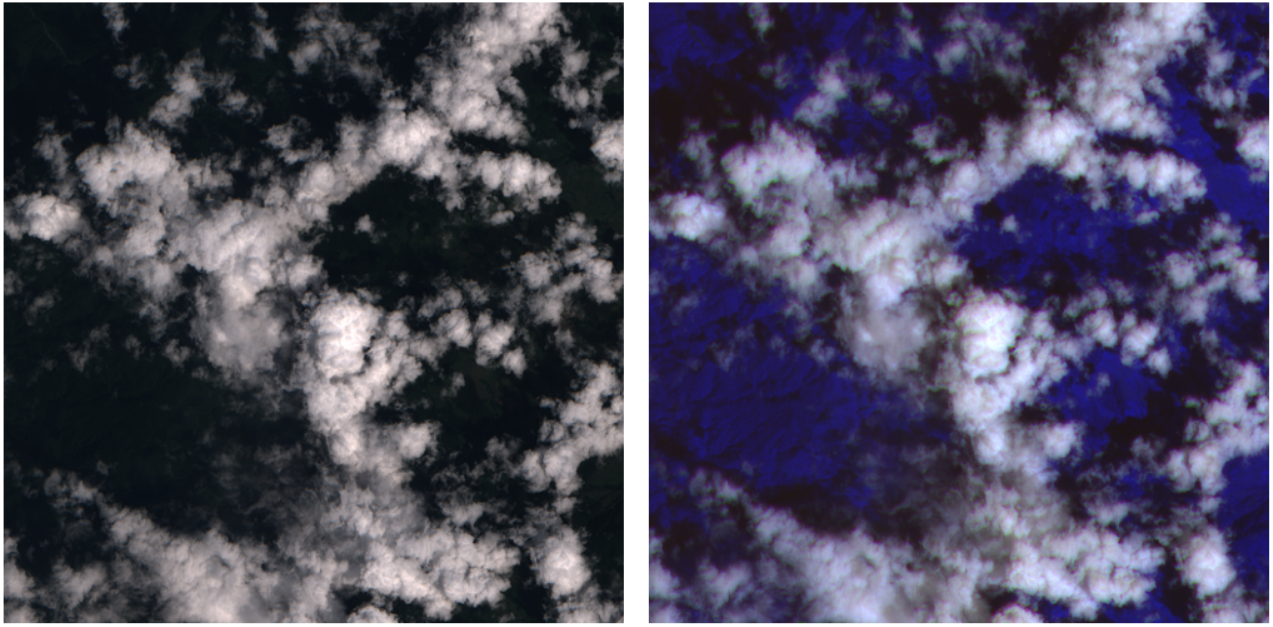}
    \caption{RGB image and false colour image (bands 1, 2, 8a) from Sentinel-2.}
\end{subfigure}
\ifarxiv
    \begin{subfigure}{\columnwidth}
\else
    \begin{subfigure}{0.8\columnwidth}
\fi
\includegraphics[width=\linewidth]{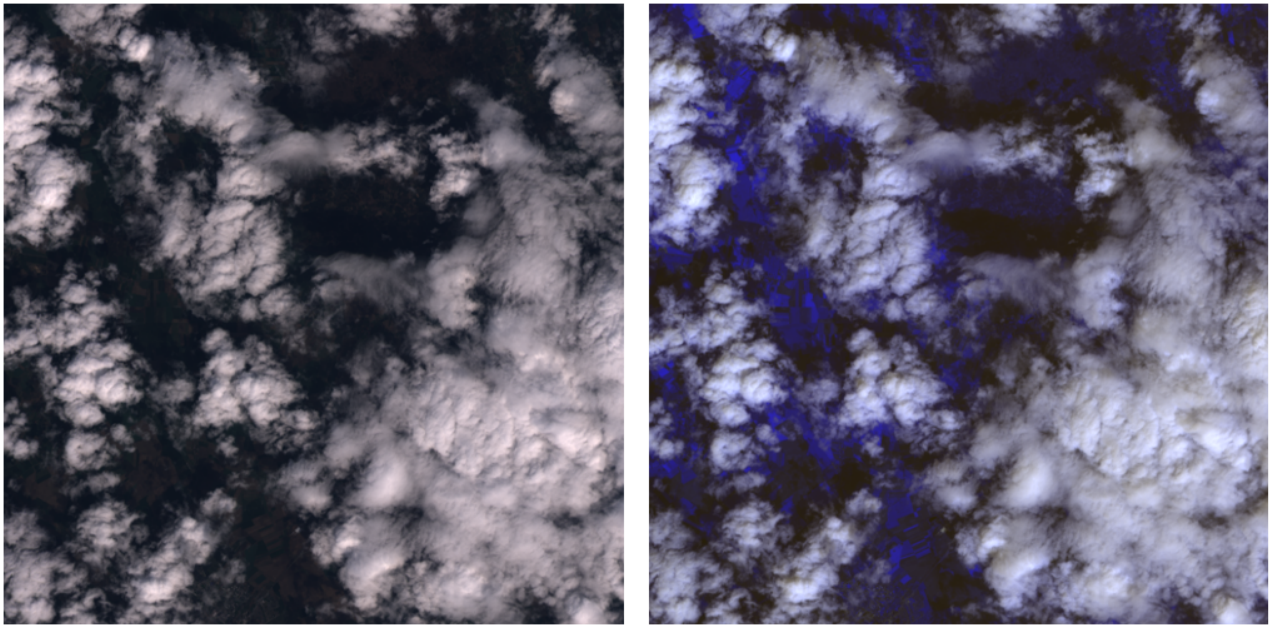}
    \caption{RGB image and false colour image (bands 1, 2, 5) from Landsat 9.}
\end{subfigure}
\caption{Hyperspectral domain gap problem. (a) Hyperspectral data from Sentinel-2~\cite{2021sentinel-2} with cloud coverage. (b) Hyperspectral data from Landsat 9~\cite{2023landsat9} in equivalent bands and with similar cloud coverage as the top row. However, a cloud detector trained on Sentinel-2 data fails to detect the presence of clouds in the data in the bottom row, indicating nontrivial differences in the distributions of data recorded by the different systems. See Sec.~\ref{sec:results:domain_gap} for quantitative results confirming the presence of hyperspectral domain gap.}
\label{fig:domain_gap}
\end{figure}

% Problem with current EO satellites - offline processing of EO data
Many EO satellites also process the data captured by the multi/hyperspectral imagers, however, this has hitherto been limited to low-level preprocessing tasks, such as data enhancement and compression. Recently, the advent of satellite-borne hardware accelerators for machine learning inference has opened up the possibility of more advanced processing. A notable example is the PhiSat-1 mission~\cite{esa-phisat-1}, which carries the HyperScout-2 payload~\cite{esposito2019in-orbit}. The payload consists of a hyperspecral imager and the Eyes of Things (EoT) ``AI on-board''~\cite{deniz2017eyes}, which executes a convolutional neural network (CNN) called CloudScout~\cite{giuffrida2020cloudscout,giuffrida2022phisat1} to perform cloud detection on the collected EO measurements. The result informs whether the geographic area in the field of view is under significant cloud cover, and only clear sky data cubes are downlinked to optimise bandwidth utilisation.

%role, with further application-driven processing typically performed further downstream at ground stations. This conventional approach introduces significant delays in converting data to actionable insights, prevents low latency coordination between end-users and space-based assets, and precludes more intelligent sensing capabilities (e.g., adaptive tasking of the sensor suite based on real-time data enhancement and analytic to improve intelligence gathering).

Generally speaking, new missions that employ new sensors (e.g., HyperScout-2) typically do not have enough representative datasets to train a CNN model. An intuitive solution is to use data captured from a previous satellite mission  to train the model (e.g., the CloudScout model was trained on data from Sentinel-2~\cite{giuffrida2020cloudscout, 2021sentinel-2}). This workaround, however, will introduce another problem called \emph{domain gap} or \emph{domain shift}~\cite{kouw2019introduction}, whereby the seemingly similar data from the training and testing domains actually differ significantly in their underlying distributions. The domain gap problem will cause the trained CNNs to perform poorly in the deployed environment; see  Fig.~\ref{fig:domain_gap} and Sec.~\ref{sec:results:domain_gap} for concrete examples. In the context of EO missions, the domain gap can be caused by several reasons:
\begin{itemize}
    \item Since hyperspectral sensors are specialised instruments manufactured in low volumes, different sensor models often differ in their effective spectral responses, spatial resolution, and signal-to-noise ratio~\cite{2021S2userhandbook, 2023L9datausershandbook, 2023hyperscout-2}. Nontrivial variations can also occur across different builds of the same sensor model due to manufacturing irregularities, sensor drifts and other physical factors~\cite{levinson2013automatic}.
    \item Hyperspectral measurements depend significantly on the ambient conditions (e.g., temperature, lighting, wind speed)~\cite{Ma2020ameliorating}, and the conditions encountered during testing may not have been recorded in the training dataset.
\end{itemize}
All the factors above collectively contribute to nontrivial differences in the data distributions. 
% Solution to problem - domain adaptation. Also mention constraints and challenges of operating in space 

Domain gap is a fundamental problem that generally affects practical applications of machine learning techniques. As a result, significant attention has been devoted to \emph{domain adaptation} methods~\cite{ganin2016domain} to counter the negative effects of domain gap (Sec.~\ref{sec:related:da} will provide a survey). However, there are major challenges to the application of domain adaptation techniques to satellite-borne machine learning:
\begin{itemize}
    \item Edge compute devices for satellite-borne machine learning are still much more limited in terms of compute capability relative to their desktop counterparts. For example, the EoT AI board~\cite{deniz2017eyes} which features an Intel Myriad 2 VPU is targeted for accelerating machine learning inference. Furthermore, the on-board CPU processor is typically catered for data acquisition and processing activities~\cite{giuffrida2022phisat1}, not for training machine learning systems or using domain adaptation techniques that are computationally costly to run.
    \item The operational constraints of a space mission, particularly limited, unreliable and/or asymmetric downlink/uplink bandwidths, lead to obstacles in data communication that affects domain adaptation, e.g., difficulties in procuring labelled target domain data and remote updating of the model deployed in space.
\end{itemize}
Sec.~\ref{sec:taskdefinition} will further discuss the challenges in the context of a concrete EO mission. Note that existing works that perform domain adaption in EO or remote sensing applications (see \autoref{tab:sda+uda+tta}) have not addressed the challenges above.

% Main contributions:
% TTA should be introduced in the paragraph below.

In this paper, we investigate domain adaptation for satellite-borne machine learning, specifically for the task of hyperspectral cloud detection. Our main contributions are:
\begin{itemize}
\item We propose novel task definitions for domain adaptation, which we named \emph{offline adaptation} and \emph{online adaptation}, that are framed in the setting of an EO mission that conducts on-board machine learning inference.
\item For offline adaptation, we propose a bandwidth-efficient supervised domain adaptation (SDA) technique that allows a satellite-borne CNN to be remotely updated while consuming only a tiny fraction of the uplink bandwidth.
\item For online adaptation, we demonstrate test-time adaptation (TTA) on a satellite-borne CNN hardware accelerator, specifically, the Ubotica CogniSAT-XE1~\cite{2023ubotica}. This shows that CNNs can be updated on realistic space hardware to account for the hyperspectral domain gap.
\end{itemize}
Our work greatly improves the viability of satellite-borne machine learning, including dealing with the inevitable problem of domain gap in hyperspecral EO applications.

%%%%%%%%%%%%%%%%%%%%%%%%%%%%%%%%%%%%%%%%%%%%%%%%%%%%%%%%%%%%%%%%%%%%%%%%%%%%%%%%%%%%%%%%%%%%%%%%
\section{Related work}\label{sec:related}
% This section reviews previous works in domain adaptation for cloud detection on EO data.
In this section, we review related works on cloud detection in EO data and on-board processing (Sec.~\ref{sec:related:hyperspectral}) and domain adaptation in remote sensing applications are also surveyed (Sec.~\ref{sec:related:da}).

\subsection{Cloud detection in EO data}\label{sec:related:hyperspectral}
% Data collected from EO satellites
EO satellites are normally equipped with multispectral or hyperspectral sensors, the main differences between the two being the spectral and spatial resolutions~\cite{madry2017electrooptical,transon2018survey}. Each ``capture'' by a multi/hyperspectral sensor produces a data cube, which consists of two spatial dimensions with as many channels as spectral bands in the sensor.

% Dealing with cloud cover in EO data
Since 66-70\% of the Earth's surface is cloud-covered at any given time~\cite{jeppesen2019cloud,li2018onboard}, dealing with clouds in EO data is essential. Two major goals are:
\begin{itemize}
\item Cloud detection, where typically the location and extent cloud coverage in a data cube is estimated;
\item Cloud removal~\cite{li2019cloud,meraner2020cloud,zi2021thin}, where the values in the spatial locations occluded by clouds are restored.
\end{itemize}
Since our work relates to the former goal, the rest of this subsection is devoted to cloud detection.

% Definition of cloud detection 
Cloud detection assigns a \emph{cloud probability} or \emph{cloud mask} to each pixel of a data cube. The former indicates the likelihood of cloudiness at each pixel, while the latter indicates discrete levels of cloudiness at each pixel~\cite{sinergise-cloud-masks}. In the extreme case, a single binary label (\emph{cloudy} or \emph{not cloudy}) is assigned to the whole data cube~\cite{giuffrida2020cloudscout}; our work focusses on this special case of cloud detection.

% Hand-crafted features vs deep features
Cloud detectors use either \emph{hand-crafted features} or \emph{deep features}. The latter category is of particular interest because the methods have shown state-of-the-art performance~\cite{lopezpuigdollers2021benchmarking, liu2021dcnet}. The deep features are extracted from data via a series of hierarchical layers in a \ac{DNN}, where the highest-level features serve as optimal inputs (in terms of some loss function) to a classifier, enabling discrimination of subtle inter-class variations and high intra-class variations~\cite{li2019deep-ieee}. The majority of cloud detectors that use deep features are based on an extension or variation of Berkeley's fully convolutional network architecture~\cite{long2015fully, shelhamer2017fully}, which was designed for pixel-wise semantic segmentation and demands nontrivial computing resources. For example, \cite{li2019deep} is based on SegNet~\cite{badrinarayanan2017segnet}, while \cite{mohajerani2018cloud, jeppesen2019cloud, yang2019cdnet, lopezpuigdollers2021benchmarking, liu2021dcnet, zhang2021cnn} are based on U-Net~\cite{ronneberger2015u-net}, all of which were not designed for on-board implementation.

% \subsection{On-board processing for cloud detection}\label{sec:related:onboard}
% Please do not write: "\cite{li2018onboard}'s" because a citation is an annotator, not a word.
On-board cloud detectors can be traced back to the thresholding-based Hyperion Cloud Cover algorithm~\cite{griffin2003cloud}, which operated on 6 of the hyperspectral bands of the EO-1 satellite. Li et al.'s on-board cloud detector~\cite{li2018onboard}
% is an integrative application of the techniques of decision tree, spectral angle map, adaptive Markov random field and dynamic stochastic resonance, 
uses hand-crafted features, but no experimental feasibility results were reported. Arguably the first \ac{DNN}-based on-board cloud detector is CloudScout~\cite{giuffrida2020cloudscout}. 
%, which operates on the HyperScout-2 imager~\cite{esposito2019in-orbit} and Eye of Things compute payload~\cite{deniz2017eyes}. As mentioned previously, CloudScout assigns a single binary label to the whole input data cube.
\autoref{tab:onboard} compares CloudScout with other more recent on-board cloud detectors. All of these detectors use the Intel Myriad 2 VPU, but none of them perform domain adaptation. Basing our work on the original CloudScout~\cite{giuffrida2020cloudscout} rather than the newer 
version~\cite{giuffrida2022phisat1}, which has lower capacity, enables us to process higher-resolution tensors and be less susceptible to adversarial attacks~\cite{du2022adversarial}.

\begin{table}[ht]
    \caption{\label{tab:onboard}Comparing on-board cloud detectors.}
    \small
    % \scriptsize
    \ifarxiv
        \begin{tabularx}{\linewidth}{p{2.36cm}p{1.9cm}X}
        \hline
        \makecell[c]{\textbf{Cloud detector}} & \makecell[c]{\textbf{Satellite}} & \makecell[c]{\textbf{\ac{DNN} characteristics}} \\
        \hline
        CloudScout~\cite{giuffrida2020cloudscout} 
            & PhiSat-1 \cite{esa-phisat-1}
            & Classifies cloudiness per image using a six-layer CNN. \\
        \hline
        CloudScout segmentation network \cite{giuffrida2022phisat1}
            & PhiSat-1 \cite{esa-phisat-1}
            & Classifies cloudiness per pixel using a variation of U-Net. \\
        \hline
        RaVAEn~\cite{ruzicka2023fast, ruzicka2022ravaen}
            & D-Orbit's ION SCV004 \cite{dorbit2023dashing}
            & Classifies cloudiness per tile of an image using a variational auto-encoder~\cite{kingma2022autoencoding} in a few-shot learning manner. \\
        \hline
        \end{tabularx}
    \else
        \begin{tabularx}{\linewidth}{p{4cm}p{2.0cm}X}
        \hline
        \makecell[c]{\textbf{Cloud detector}} & \makecell[c]{\textbf{Satellite}} & \makecell[c]{\textbf{\ac{DNN} characteristics}} \\
        \hline
        CloudScout~\cite{giuffrida2020cloudscout} 
            & PhiSat-1 \cite{esa-phisat-1}
            & Classifies cloudiness per image using a six-layer CNN. \\
        \hline
        CloudScout segmentation network \cite{giuffrida2022phisat1}
            & PhiSat-1 \cite{esa-phisat-1}
            & Classifies cloudiness per pixel using a variation of U-Net. \\
        \hline
        RaVAEn~\cite{ruzicka2023fast, ruzicka2022ravaen}
            & D-Orbit's ION SCV004 \cite{dorbit2023dashing}
            & Classifies cloudiness per tile of an image using a variational auto-encoder~\cite{kingma2022autoencoding} in a few-shot learning manner. \\
        \hline
        \end{tabularx}        
    \fi
\end{table}

% \begin{table}[ht]
%     \caption{\label{tab:onboard}Comparing on-board cloud detectors.}
%     \small
%     % \scriptsize
%     % \begin{tabularx}{\linewidth}{|p{2.36cm}|p{1.9cm}|X|}
%     \begin{tabularx}{\linewidth}{|p{3.1cm}|p{2.0cm}|X|}
%     \hline
%     \makecell[c]{\textbf{Cloud detector}} & \makecell[c]{\textbf{Satellite}} & \makecell[c]{\textbf{\ac{DNN} characteristics}} \\
%     \hline
%     CloudScout~\cite{giuffrida2020cloudscout}
%         & \multirow{4}{2cm}{PhiSat-1 \cite{esa-phisat-1}}
%         & Classifies cloudiness per image using a six-layer CNN. \\
%     \cline{1-1}\cline{3-3}
%     CloudScout seg- mentation net \cite{giuffrida2022phisat1}
%         & 
%         & Classifies cloudiness per pixel using a variation of U-Net. \\
%     \hline
%     RaVAEn~\cite{ruzicka2023fast, ruzicka2022ravaen}
%         & D-Orbit's ION SCV004 \cite{dorbit2023dashing}
%         & Classifies cloudiness per tile of an image using a variational auto-encoder~\cite{kingma2022autoencoding} in a few-shot learning manner. \\
%     \hline
%     \end{tabularx}
% \end{table}

Relevant to on-board processing but not cloud detection, Mateo-Garcia et al.~\cite{mateogarcia2023inorbit} experimented with histogram matching but settled on offline retraining for supervised domain adaptation (see Sec.~\ref{sec:related:da}).

\subsection{Domain adaptation in remote sensing applications}\label{sec:related:da}
% Definition of domain adaptation
% Domain adaptation are methods that update the model to work in the target domain...
% Domain shift = distribution shift between a set of source-domain/training data and a set of target-domain/test data~\cite{zhou2022}

%\hlc[green]{Dzung:How about we split this section into two subsections: offline and online domain adaptations?}

%\hlc[green]{Dzung: In offline adaptation, we review recent domain adaptation papers for cloud detection but no need to distinguish UDA and SDA. Then, we argue that all existing offline adaptation methods will update ALL parameters which are BANDWITH-INEFFICIENT. Then we show our method just update FEW important parameters, which are BANDWITH-EFFICIENT $=>$ our paper is novel in that way}

%\hlc[green]{Dzung: In online adaptation: we could review TTA for cloud detection or even TTA for image classification. But we emphasise none of existing TTA methods are really implemented in edge device, Then we claim we are the first to actually implement TTA to edge device $=>$ our paper is novel in that way}

Domain generalisation refers to learning of invariant representations using data from multiple source domains to achieve generalisation to any out-of-distribution data in the target domain~\cite{zou2022learning, liang2023comprehensive}. When target data becomes available, domain adaptation rather than domain generalisation can be used. As recent surveys~\cite{farahani2021brief, liu2022deep, peng2022domain, zhang2022transfer, fang2023source, liang2023comprehensive, singhal2023domain, yu2023comprehensive} reveal, there is a wide variety of domain adaptation methods, but the two main types that have been applied to remote sensing thus far are \ac{SDA} and \ac{UDA}.
% These methods can be categorized into three different adaptation settings:

% \subsubsection{\Ac{SDA}}\label{sec:sda}

\Ac{SDA} methods use labelled data in both the source and target domains~\cite{kellenberger2021deep}, although the quantity of labelled data in the target domain is typically smaller.
% Common techniques such as data augmentation and ensemble learning serve as the algorithmic building blocks.
Algorithmic building blocks include fine-tuning, data augmentation and ensemble learning~\cite{zhou2023domain}.
\autoref{tab:sda+uda+tta} compares sample applications of \ac{SDA} to multispectral image classification.
% one of which \cite{mateogarcia2020transferring} focusses on cloud detection.
\Ac{SSDA} targets the scenario where there is a small amount of labelled data but a good amount of unlabelled data in the target domain~\cite{liu2022deep}. The appeal of \ac{SSDA} wanes~\cite{lucas2023bayesian} as \ac{UDA} rapidly advances.

\begin{table*}[ht]
    \caption{\label{tab:sda+uda+tta}Applications of \ac{SDA} and \ac{UDA} to multispectral/hyperspectral image classification. References with an asterisk (${}^\ast$) are specifically about cloud detection.}
    \small
    \begin{tabularx}{\linewidth}{l|lllX}
        \hline
        & \textbf{Ref.} & \textbf{Source dom.} & \textbf{Target dom.} & \makecell[c]{\textbf{Salient characteristics}} \\
        \hline
        \multirow{4}{*}{\rotatebox[origin=c]{90}{SDA}}
            & \cite{shendryk2019deep}
            & PlanetScope
            & Sentinel-2
            & An ensemble of three CNN models is pre-trained on the source data, and fine-tuned on the target data. \\
        \cline{2-5}
            & \cite{mateogarcia2020transferring}$^\ast$
            & Landsat-8
            & Proba-V
            & A U-Net-based CNN is trained on the source data and three images from the target domain. \\
        \cline{2-5}
            & \cite{mateogarcia2023inorbit}
            & Sentinel-2
            & \makecell[lt]{D-Sense\\images}
            & A CNN is trained on the source data and four images from the target domain. Model retraining happens on ground and updated model is uplinked to the satellite. \\
        \hline
        \multirow{3}{*}{\rotatebox[origin=c]{90}{UDA}}
            & \cite{segalrozenhaimer2020cloud}$^\ast$
            & WorldView-2
            & Sentinel-2
            & A DeepLab-like \cite{chen2018deeplab} CNN is trained on the source data, and adapted to the target domain through a \ac{DANN} \cite{ganin2016domain}. \\
        \cline{2-5}
            & \cite{mateogarcia2021cross}$^\ast$
            & Landsat-8
            & Proba-V
            & A five-layer fully connected neural network is trained on an upscaled version of the source data, and adapted to the target domain through \emph{generative domain mapping}~\cite{liu2022deep}, where a cycle-consistent generative adversarial network~\cite{zhu2017unpaired} maps target data to the upscaled source domain. \\
        \hline
        \multirow{2}{*}{\rotatebox[origin=c]{90}{TTA}}
            & \cite{xu2022source}
            & Dioni
            & \makecell[lt]{HyRANK,\\Pavia}
            & A 3D-CNN \cite{li2017spectral} is trained on the source data and adapted to the target domain through Contrastive Prototype Generation and Adaptation~\cite{qiu2021source}. \\
        \hline
    \end{tabularx}
\end{table*}

% \subsubsection{\Ac{UDA}}\label{sec:uda}

\Ac{UDA} methods use labelled data in the source domain but only unlabelled data in the target domain~\cite{kellenberger2021deep}. \ac{UDA} methods based on deep learning can automate learning of transferable features, and can be used in two broad scenarios~\cite{fang2023source}:
\begin{enumerate}
    \itemsep0.7em 
    \item When source data is available for adapting the model to the target domain, this type of \ac{UDA} is so-called conventional.

    \item When source data is unavailable but a source model is available to be adapted to the target domain, this type of \ac{UDA} is called source-free (\emph{unsupervised}) domain adaptation (SFDA), or equivalently \ac{TTA}. In theory, source-free \emph{supervised} domain adaptation is feasible but practically meaningless. \ac{TTA} is further discussed in Sec.~\ref{sec:tta}.
    
    % \item When source data is unavailable but a source model with accessible parameters is available to be adapted to the target domain, this type of \ac{UDA} is called source-free (\emph{unsupervised}) domain adaptation (SFDA). In theory, source-free \emph{supervised} domain adaptation is feasible but practically meaningless.
\end{enumerate}
A common technique employed by conventional \ac{UDA} schemes is \emph{alignment}, i.e., transforming either raw inputs or features such that the resultant probability distributions (marginal or/and conditional) in the source and target domains are as close as possible~\cite{kouw2019introduction, liu2021adversarial, zhou2023domain}. The closeness of distributions can be quantified with a divergence measure, e.g., Kullback-Leibler divergence~\cite{kullback1951information}.
% Kullback-Leibler divergence~\cite{kullback1951information}, \ac{MMD}~\cite{gretton2006kernel}, $\mathcal{H}$-divergence~\cite{bendavid2010theory}, \ac{CORAL}~\cite{sun2016return}, Jensen-Shannon divergence~\cite{zhao2019learning}.
At least two classes of deep, conventional \ac{UDA} methods are discernible~\cite{farahani2021brief, peng2022domain}:

%\hlc[green]{Dzung: I think both ``Discrepancy-based" and "Adversarial learning" aim to minimise the discrepancy of two domains in feature space, so should we split them like they are two separate approaches?}
\begin{enumerate}
    \itemsep0.7em 
    \item \textbf{Discrepancy-based} methods perform \emph{statistical divergence alignment} \cite{liu2022deep}, i.e., match marginal or/and conditional distributions between domains by integrating into a \ac{DNN} adaptation layers designed to minimise domain discrepancy in a latent feature space.
    % For example, TCANet~\cite{garea2019tcanet} uses transfer component analysis~\cite{pan2011domain} to extract features that minimises the \ac{MMD} between source- and target-domain distributions.
    See \cite[Sec. VI.A]{peng2022domain} for a survey of applications of discrepancy-based \ac{UDA} to EO image classification.
    % Examples include \cite{garea2019tcanet, wang2019domain, li2020two}.
 
    \item \textbf{Adversarial-learning} methods learn transferable and domain-invariant features through adversarial learning. A well-known method is using a \ac{DANN} \cite{ganin2016domain}, which comprises a feature extractor network connected to a label predictor and a domain classifier. Training the network parameters to (i) minimise the loss of the label predictor but (ii) maximise the loss of the domain classifier, promotes the emergence of domain-invariant features. 
    \autoref{tab:sda+uda+tta} compares sample applications of \ac{UDA} to cloud detection.
    See \cite[Sec. VI.B]{peng2022domain} for a survey of applications of adversarial-learning \ac{UDA} to EO image classification.
    % Examples include \cite{cheng2023soft, fang2022confident, wang2022hyperspectral, deng2020deep, segalrozenhaimer2020cloud}.    
\end{enumerate}
None of the methods covered so far are applicable when (i) target data is unavailable during training, and (ii) source data is unavailable during knowledge transfer. Instead, \ac{TTA} becomes necessary.

\subsubsection{\Ac{TTA}}\label{sec:tta}

\Ac{TTA} is \ac{UDA} without access to the source data. Multiple classifications \cite{fang2023source, liang2023comprehensive, yu2023comprehensive} of \ac{TTA} methods exist but the types of interest here are \emph{white-box} (where model parameters are accessible for adaptation) and \emph{online} (where unlabeled target data is ingested in a stream and processed once). Examples of white-box online \ac{TTA} include:
\begin{itemize}
    \item Test Entropy Minimisation (Tent)~\cite{wang2021tent}: This method adapts a probabilistic and differentiable model by minimising the Shannon entropy of its predictions. For each \ac{BN}~\cite{ioffe2015batch} layer in a \ac{DNN}, Tent updates (i) the normalisation statistics $\mu,\sigma$ in the forward pass, and (ii) the affine transformation parameters $\gamma,\beta$ in the backward pass. See Sec.~\ref{sec:online_adapt:tent} for more details. 

    \item \Ac{DUA}~\cite{mirza2022norm}: This method modulates the ``momentum" of \ac{BN} layers with a decay parameter, which helps stabilise the adaptation process. See Sec.~\ref{sec:online_adapt:dua} for more details. \Ac{DUA} shows similar adaptation performance to Tent~\cite[Sec. 4]{mirza2022norm}. 
\end{itemize}
There is a lack of reported works on  applying \ac{TTA} to EO applications; this shortage accentuates the novelty of our work. \autoref{tab:sda+uda+tta} includes one sample application of \ac{TTA} to hyperspectral image classification.

\ifarxiv
    \vspace{-2ex}
\fi
\section{Domain adaptation tasks for EO mission}\label{sec:taskdefinition}
In this section, we describe two domain adaptation tasks---offline (Sec.~\ref{sec:offline_adaptation}) and online adaptation (Sec.~\ref{sec:online_adaptation}) ---for satellite-borne machine learning applications. The significance of our formulations derives from framing the formulations in the context of a concrete EO mission that has successfully demonstrated an onboard machine learning task. We thus begin by describing the mission context (Sec.~\ref{sec:missioncontext}), before defining the domain adaptation tasks.

\subsection{Cloud detection on PhiSat-1}\label{sec:missioncontext}
The aim of the PhiSat-1 nanosatellite mission is to demonstrate the feasibility and usefulness in bringing AI on board a satellite \cite{esa-phisat-1}. It involves the use of a CNN called CloudScout~\cite{giuffrida2020cloudscout} to perform cloud detection on data cubes captured by the hyperspectral imaging payload. More formally: preprocessing is first performed on the output of the hyperspectral imager (e.g., radiometric and geometric corrections, stacking and alignment, as well as band selection and normalisation) to yield a data cube $x$. The CNN-based cloud detector can be formalised as the function $y = f(x ; \theta)$ such that the assigned label
\begin{align}
y = \begin{cases}
        1 & \text{if $x$ contains significant cloud coverage;} \\
        0 & \text{otherwise}.
    \end{cases}
\end{align}
In the case where $y = 1$, $x$ is discarded (precluded from being transmitted to ground). The weights $\theta$ define the function implemented by $f$; details of the CNN architecture will be provided in Sec.~\ref{sec:building}. In PhiSat-1, the CNN is executed on the EoT AI board, particularly the embedded Intel Myriad 2 VPU which was experimentally proven to be able to withstand the harshness of the space environment~\cite{furano2020towards}.

% As described in~\cite{giuffrida2020cloudscout}, the main performance and design requirements on the CNN are as follows:
% \begin{itemize}
%     \item Minimum accuracy of 85\% in cloud detection to enable reasonable results even in uncommon situations, e.g., clouds on ice, or clouds on salt-lake.
    
%     \item Maximum \ac{FP} rate of 1.2\% to avoid the loss of potentially good hyperspectral data.
    
%     \item Maximum memory footprint of 5 MB to permit the model (specifically, the weights $\theta$) to be remotely updated under the uplink bandwidth limit (\hl{is there a number from cloudscout paper?}).

% \end{itemize}
% The third requirement is due to bandwidth asymmetry in satellite communications~\cite{papadimitriou2007tcp}, where typically larger bandwidths are allocated for downlinking to support large-volume telemetry, while much smaller bandwidths are allocated for uplinking since telecommand traffic is sparser. Details of the CNN architecture that meet the requirements will be provided in Sec.~\ref{sec:building}. 
    
\subsection{Pre-deployment model training}\label{sec:pre-deployment}

\begin{figure}[ht]
\centering
\includegraphics[width=\linewidth]{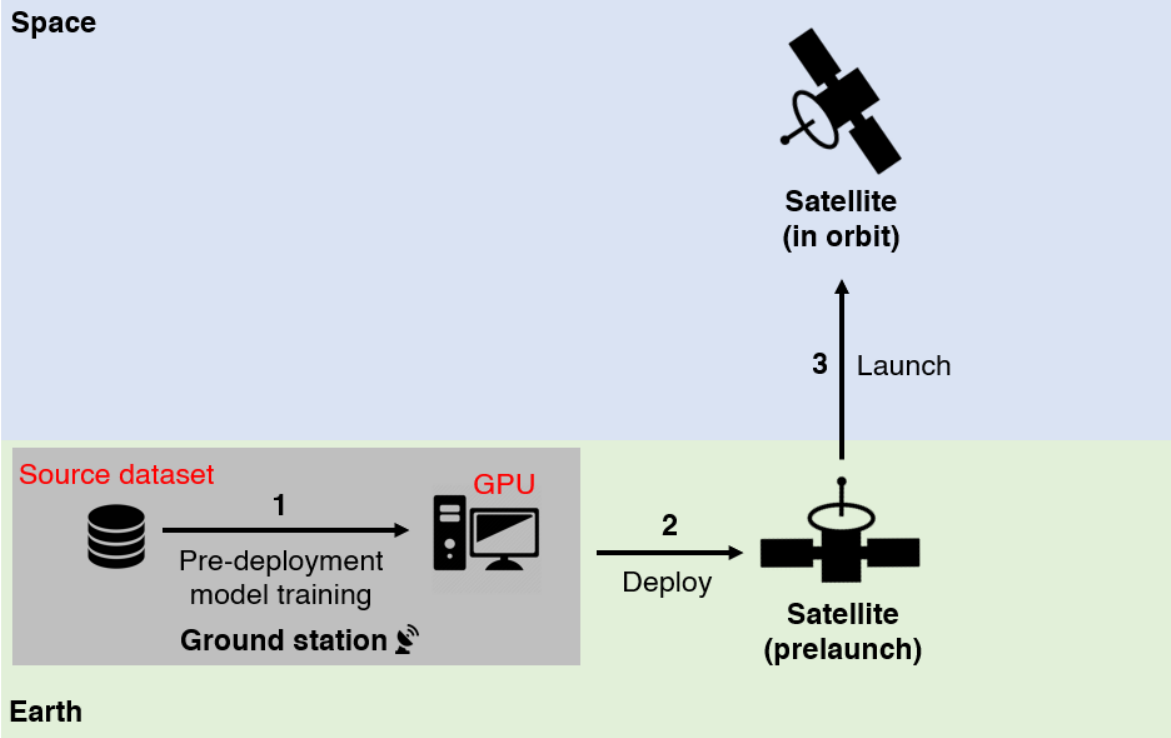}
\caption{Pre-deployment model training, satellite deployment, and launch.}
\label{fig:offline_training}
\end{figure}

Prior to deployment and launch, $f(\cdot\,;\,\theta)$ is trained on a labelled dataset $\cD^s = \{ x^s_i, y^s_i \}_{i=1}^{N^s}$, where each $x^s_i$ is a preprocessed data cube and $y^{s}_i$ is the ground truth label. $\cD^s$ is called the \textbf{source dataset} as it is collected from a relevant \textbf{source domain}, e.g., from a previous EO mission. Details on building $\cD^s$ and training $f(\cdot\,;\,\theta)$ will be provided in Sec.~\ref{sec:dataset_construction} and \ref{sec:building} respectively. Note that the training is conducted on ground, e.g., on a GPU workstation. Once trained, we obtain a \textbf{source cloud detector} $f(\cdot\,;\,\theta^s)$ parameterised by \textbf{source weights} $\theta^s$, which is then deployed onto the satellite and launched into orbit. We distinguish between two copies of the model: $f^d(\cdot\,;\,\theta^s)$ and $f^g(\cdot\,;\,\theta^s)$, i.e., the deployed and ground versions. Both versions are identical at the time of deployment. The steps from pre-deployment model training to launch are depicted in Fig.~\ref{fig:offline_training}.

\subsection{Post-deployment domain adaptation}\label{sec:post_deployment}

Due to domain gap, it is expected that $f^d(\cdot\,;\,\theta^s)$ will not be accurate when applied on the data collected in orbit, i.e., data in the \textbf{target domain}. Thus, it is necessary to perform domain adaptation on $f^d(\cdot\,;\,\theta^s)$ to obtain a \textbf{target cloud detector} $f^d(\cdot\,;\,\theta^t)$ parameterised by \textbf{target weights} $\theta^t$. To this end, a new unlabelled dataset $\tilde{\cD}^t = \{ x^t_j\}_{j=1}^{N^t}$ (called the \textbf{unlabelled target dataset}) is collected onboard.

\subsubsection{Offline adaptation}\label{sec:offline_adaptation}
\begin{figure}[h]
\centering
\includegraphics[width=\linewidth]{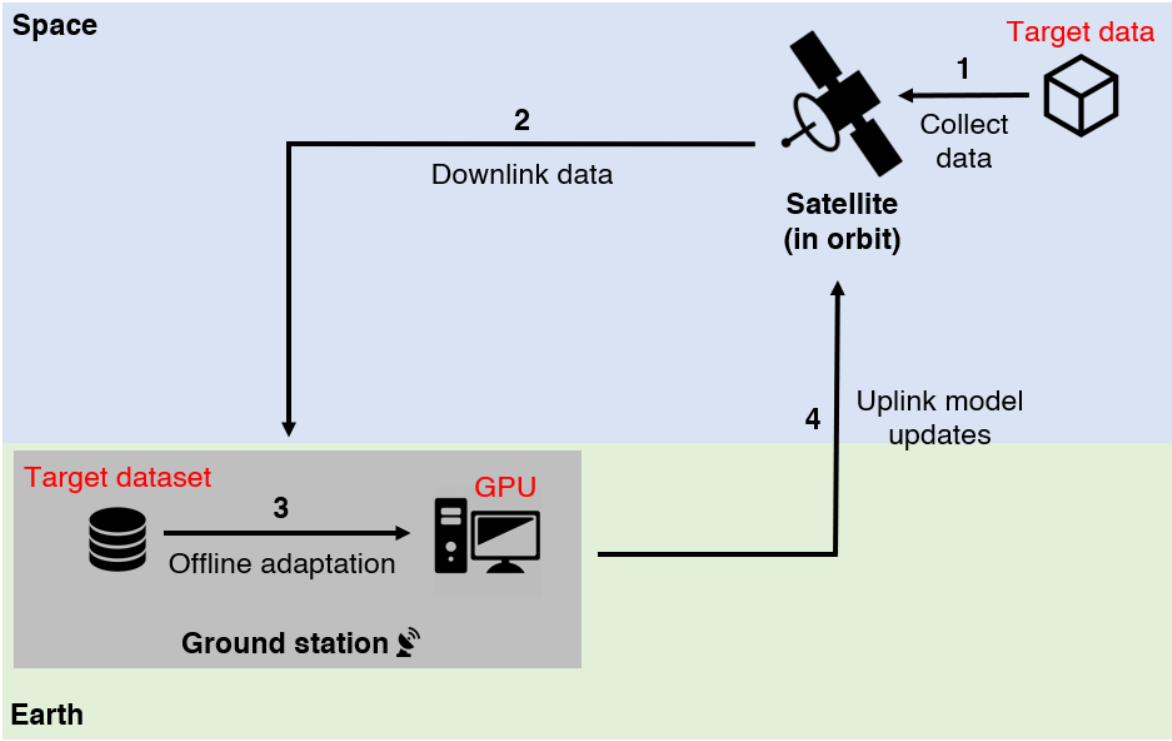}
\caption{Collection and downlinking of target data, offline adaptation performed on ground station, and uplinking model updates.}
\label{fig:offline_adaptation}
\end{figure}

% describe the situation of offline adaptation and a naive solution 
Offline adaptation assumes the ability to downlink $\tilde{\cD}^t$. We can thus label $\tilde{\cD}^t$ (e.g., via manual labelling) and form the \textbf{labelled target dataset} $\cD^t = \{ x^t_j, y^t_j \}_{j=1}^{N^t}$. Details on building $\cD^t$ will be provided in Sec.~\ref{sec:dataset_construction}. A straightforward approach to domain adaptation is to update $f^g$ on $\cD^t$ to obtain $\theta^t$, which amounts to conducting SDA (see Sec.~\ref{sec:related:da}). Then, $f^d$ is updated by uplinking $\theta^t$ to remotely replace $\theta^s$. The steps from downlinking $\tilde{\cD}^t$ to uplinking $\theta^t$ are depicted in Fig.~\ref{fig:offline_adaptation}
% To update $f^d$, the target weights $\theta^t$ is uplinked to remotely replace source weights $\theta^s$ of $f^d$ \hlc[green]{with $\theta^t$}. Offline adaptation is depicted in Fig.~\ref{fig:offline_adaptation}.

% problem with uplink bandwidth
However, the ability to downlink $\tilde{\cD}^t$ does not imply the ability to uplink $\theta$, particularly if the architecture of $f$ is complex; for example, ResNet50 with $\approx$ 23 M single-precision floating-point (FP32) weights has a memory footprint of 94.37 MB. This is because many satellite communication bandwidths are asymmetric~\cite{papadimitriou2007tcp}, in that larger bandwidths are allocated for downlinking to support large-volume telemetry, while much smaller bandwidths are allocated for uplinking since telecommand traffic is sparser. Indeed, PhiSat-1 restricts the maximum memory footprint of $f$ to 5~MB to permit the model to be remotely updated~\cite{giuffrida2020cloudscout}. However, this limits the learning capacity of $f$, which could reduce its accuracy and increase its susceptibility to adversarial attacks~\cite{du2022adversarial}.

% solution to problem - FISH Mask
To alleviate the uplink restrictions on model size, it is vital to perform SDA in a bandwidth-efficient manner. This can be achieved by restricting the number of individual weights of $\theta^s$ that are changed when updating $f^g$ on $\cD^t$. Then, only a small number of refined weights are uplinked to remotely update $f^d$. The problem is summarised as follows:
\begin{mybox}{Problem 1: Bandwidth-efficient SDA}
Given labelled target dataset $\cD^t = \{ x^t_j, y^t_j \}_{j=1}^{N^t}$ and source cloud detector $f^g(\cdot\,;\,\theta^s)$, update $f^g$ using $\cD^t$ by making as few changes to the source weights $\theta^s$ as possible.
\end{mybox}

In Sec.~\ref{sec:offline_sda}, we will describe a solution to Problem 1 that enables only a small fraction of the weights to be updated without noticeable impacts to cloud detection accuracy. This enables large models to be used and remotely updated through the thin uplink channel.

\subsubsection{Online adaptation}\label{sec:online_adaptation}
% describe the situation of online adaptation
Online adaptation directly updates $f^d$ on $\tilde{\cD}^t$ onboard the satellite. Therefore, it does not require downlinking $\tilde{\cD}^t$ to ground. The source dataset $\cD^s$ is also assumed to be unavailable on the satellite, due to lack of storage.  Hence, the problem is an instance of TTA (see Sec.~\ref{sec:tta}). 

An important requirement of online adaptation is a suitable runtime environment on satellite-borne edge compute devices that can execute the TTA algorithm. A runtime environment that stores a full-fledged machine learning framework (e.g., PyTorch~\cite{2023pytorch}) and its associated dependencies can require up to several gigabytes of disk space. Such resources are not available on edge devices. Furthermore, the runtime environment may need to be updated during the life of the mission due to bug patches. Uplinking these updates may also not be possible especially for large runtime environments. The problem is summarised as follows: 
\begin{mybox}{Problem 2: TTA on satellite hardware}
Given unlabelled target dataset $\tilde{\cD}^t = \{ x^t_j \}_{j=1}^{N^t}$ and source cloud detector $f^d(\cdot\,;\,\theta^s)$, update $f^d$ using $\tilde{\cD}^t$ in a runtime environment suitable for satellite-borne edge compute hardware.
\end{mybox}

In Sec.~\ref{sec:online_tta}, we will describe our steps to execute state-of-the-art TTA algorithms on a testbed that simulates the compute payload of a EO satellite. This establishes the viability of TTA on space hardware.

\ifarxiv
    \vspace{-2ex}
\fi
\section{Dataset construction}\label{sec:dataset_construction}
In this section, we provide details of constructing the labelled source dataset $\cD^s$ and labelled target dataset $\cD^t$.

% \hl{Put both S2 and L9 here. No need to indicate which one is source or target at this stage.}

\subsection{Sentinel-2}\label{sec:sentinel2}
The Sentinel-2 Cloud Mask Catalogue \cite{francis_alistair_2020_4172871} contains cloud masks for 513 Sentinel-2A Top-of-Atmosphere (TOA) reflectance \cite{2021sentinel-2} data cubes (1024$\times$1024 pixels) collected from a variety of geographical regions. Each data cube has 13 spectral bands with a spatial resolution of 20 m. Following \cite{giuffrida2020cloudscout}, we spatially divided the data cubes into 2052 data (sub)cubes of 512$\times$512 pixels each. 

% Ground truth (binary) labels were obtain by thresholding the number of cloudy pixels in the cloud masks. We followed \cite{giuffrida2020cloudscout} by applying a threshold of 30\% resulting to \hl{dataset $\cD^s_\text{TH30}$, and 70\% resulting to dataset $\cD^s_\text{TH70}$}. The reason for producing the two variants of $\cD^s$ will be provided in Sec.~\ref{sec:source_detector}. Each dataset was further divided into training and testing sets. 

% In Sec.~\ref{sec:results:domain_gap}, \hl{where we will establish the presence of domain gap, we will also employ these datasets as ``target'' datasets.}

\subsection{Landsat 9}\label{sec:landsat9}
% In a real-world EO mission, the target data is collected onboard a satellite in orbit. We instead used Landsat 9 data products from USGS Earth Explorer~\cite{2023usgs} to simulate the collection of the target dataset. These data products were selected due to its similarities with Sentinel-2. As shown in Tab.~\ref{tab:spectral-bands}, there are 8 bands that closely overlap one another in terms of their central wavelength (CW), bandwidth (BW), and spatial resolution (SR). They contain data cubes, each with 11 spectral bands and a spatial resolution of 15~m, 30~m, and 100~m. 

Landsat 9 data products from USGS Earth Explorer~\cite{2023usgs} were selected due to there similarities with Sentinel-2. As shown in \autoref{tab:spectral-bands}, there are 8 bands that closely overlap with one another in terms of their central wavelength (CW), bandwidth (BW), and spatial resolution (SR). These data products contain data cubes, each with 11 spectral bands and a spatial resolution of 15~m, 30~m or 100~m. Cloud masks were also provided. 

The data cubes were preprocessed in a similar manner as in \cite{francis_alistair_2020_4172871} by (i) converting the quantised and calibrated scaled Digital Numbers to TOA reflectances, and (ii) resampling bands to a spatial resolution of 30 m using bilinear interpolation. We also spatially divided the data cubes into 2000 data (sub)cubes of 512×512 pixels each. 

% Cloud confidence masks were also provided in the data products which were used to create ground truth labels by converting the confidence masks to binary masks, and then thresholding the binary mask with 30\% and 70\% cloudiness resulting to \hl{datasets $\cD^t_\text{TH30}$ and $\cD^t_\text{TH70}$} respectively. Each dataset was further divided into training and testing sets. In Sec.~\ref{sec:adapting}, $\cD^t_\text{TH70}$ is used to adapt the cloud detectors to the target domain. 

\subsection{Ground-truth labels and their usage}\label{sec:groundtruths}
For the source domain, if Sentinel-2 data are used, then Landsat 9 data are used for the target domain. Likewise, if Landsat 9 data are used for the source domain, then Sentinel-2 data are used for the target domain.

To train the source cloud detector (Sec.~\ref{sec:pre-deployment}), source data cubes were assigned a binary label (\emph{cloudy} vs. \emph{not cloudy}) by thresholding the number of cloudy pixels in the cloud masks. We followed \cite{giuffrida2020cloudscout} by applying thresholds of 30\% and 70\% to produce labelled source datasets $\cD^s_\text{TH30}$ and $\cD^s_\text{TH70}$ respectively. Each of $\cD^s_\text{TH30}$ and $\cD^s_\text{TH70}$ was further divided into training and testing sets.

To adapt the source cloud detector in the offline setting (Sec.~\ref{sec:offline_adaptation}),  target data cubes were assigned a binary label by applying a 70\% cloudiness threshold on the cloud masks to produce a labelled target dataset $\cD^t_\text{TH70}$. This dataset was further divided into training and testing sets. Recall in the online setting (Sec.~\ref{sec:online_adaptation}), an unlabelled target dataset $\tilde{\cD}^t$ is only required for adaptation.

\begin{table*}[h]
    \centering
    % \scriptsize
    \small
    \caption{Spectral bands of Sentinel-2 and Landsat 9. Text in blue are the 8 bands that closely overlap with one another in terms of their central wavelength (CW), bandwidth (BW), and spatial resolution (SR).}
    \label{tab:spectral-bands}
    % \begin{tabular}{|p{4.0cm}|p{1.5cm}|p{1.5cm}|p{1.5cm}|p{4.0cm}|p{1.5cm}|p{1.5cm}|p{1.5cm}|}
    \begin{tabularx}{\linewidth}{lRRRllRRR}
        \hline
        % \rowcolor{black}
        \multicolumn{4}{c}{\Gape{\makecell[t]{\textcolor{black}{\textbf{Senitinel-2 (13 bands)}}}}} & & \multicolumn{4}{c}{\Gape{\makecell[t]{\textcolor{black}{\textbf{Landsat 9 (11 bands)}}}}} \\
        % \hline
        \textbf{Spectral bands} & \makecell{\textbf{CW} \\ \textbf{(nm)}} & \makecell{\textbf{BW} \\ \textbf{(nm)}} & \makecell{\textbf{SR} \\ \textbf{(m)}} & \textcolor{white}{AAA} & \textbf{Spectral bands} & \makecell{\textbf{CW} \\ \textbf{(nm)}} & \makecell{\textbf{BW} \\ \textbf{(nm)}} & \makecell{\textbf{SR} \\ \textbf{(m)}} \\
        \hline
        \textcolor{blue}{B01 - Coastal Aerosol} & \textcolor{blue}{442.7} & \textcolor{blue}{21} & \textcolor{blue}{60} & & \textcolor{blue}{B01 - Coastal Aerosol} & \textcolor{blue}{443} & \textcolor{blue}{16} & \textcolor{blue}{30} \\
        % \hline
        \textcolor{blue}{B02 - Blue} & \textcolor{blue}{492.4} & \textcolor{blue}{66} & \textcolor{blue}{10} & & \textcolor{blue}{B02 - Blue} & \textcolor{blue}{482} & \textcolor{blue}{60} & \textcolor{blue}{30} \\
        % \hline
        \textcolor{blue}{B03 - Green} & \textcolor{blue}{559.8} & \textcolor{blue}{36} & \textcolor{blue}{10} & &\textcolor{blue}{B03 - Green} & \textcolor{blue}{561.5} & \textcolor{blue}{57} & \textcolor{blue}{30} \\
        % \hline
          &  &  &  &  & B08 - Panchromatic & 589.5 & 173 & 15\\
        % \hline
        \textcolor{blue}{B04 - Red} & \textcolor{blue}{664.6} & \textcolor{blue}{31} & \textcolor{blue}{10} & & \textcolor{blue}{B04 - Red} & \textcolor{blue}{654.5} & \textcolor{blue}{37} & \textcolor{blue}{30}\\
        % \hline
        B05 - Red Edge 1 & 704.1 & 15 & 20 & & & & & \\
        % \hline
        B06 - Red Edge 2 & 740.5 & 15 & 20 & & & & & \\
        % \hline
        B07 - Red Edge 3 & 782.8 & 20 & 20 & & & & & \\
        % \hline
        B08 - NIR & 832.8 & 106 & 10 & & & & & \\
        % \hline
        \textcolor{blue}{B08A - Narrow NIR} & \textcolor{blue}{864.7} & \textcolor{blue}{21} & \textcolor{blue}{20} & & \textcolor{blue}{B05 - NIR} & \textcolor{blue}{865} & \textcolor{blue}{28} & \textcolor{blue}{30} \\
        % \hline
        B09 - Water Vapour & 945.1 & 20 & 60 & & & & & \\
        % \hline
        \textcolor{blue}{B10 - SWIR - Cirrus} & \textcolor{blue}{1373.5} & \textcolor{blue}{31} & \textcolor{blue}{60} & & \textcolor{blue}{B09 - Cirrus} & \textcolor{blue}{1373.5} & \textcolor{blue}{21} & \textcolor{blue}{30} \\
        % \hline
        \textcolor{blue}{B11 - SWIR 1} & \textcolor{blue}{1613.7} & \textcolor{blue}{91} & \textcolor{blue}{20} & & \textcolor{blue}{B06 - SWIR 1} & \textcolor{blue}{1608.5} & \textcolor{blue}{85} & \textcolor{blue}{30} \\
        % \hline 
        \textcolor{blue}{B12 - SWIR 2} & \textcolor{blue}{2202.4} & \textcolor{blue}{175} & \textcolor{blue}{20} & & \textcolor{blue}{B07 - SWIR 2} & \textcolor{blue}{2200.5} & \textcolor{blue}{187} & \textcolor{blue}{30} \\
        % \hline 
          &  &  &  &  & B10 - Thermal & 10895 & 590 & 100 \\
        % \hline 
          &  &  &  &  & B11 - Thermal & 12005 & 1010 & 100 \\
        \hline
    \end{tabularx}
\end{table*}

%%%%%%%%%%%%%%%%%%%%%%%%%%%%%%%%%%%%%%%%%%%%%%%%%%%%%%%%%%%%%%%%%%%%%%%%%%%%%%%%%%%%%%%%%%%%%%%%
\ifarxiv
    \vspace{-2ex}
\fi
\section{Building the cloud detector}\label{sec:building}
In this section, we provide details of the steps involved in the pre-deployment model training stage (Sec.~\ref{sec:pre-deployment}). More specifically, we describe the CNN architectures used for on-board cloud detection (Sec.~\ref{sec:architecture}) and the training procedure for the source cloud detector $f(\cdot\,;\,\theta^s)$ (Sec.~\ref{sec:source_detector}). 

\subsection{CNN architectures for cloud detection}\label{sec:architecture} 
\begin{figure}[ht]
\centering
\includegraphics[width=\linewidth]{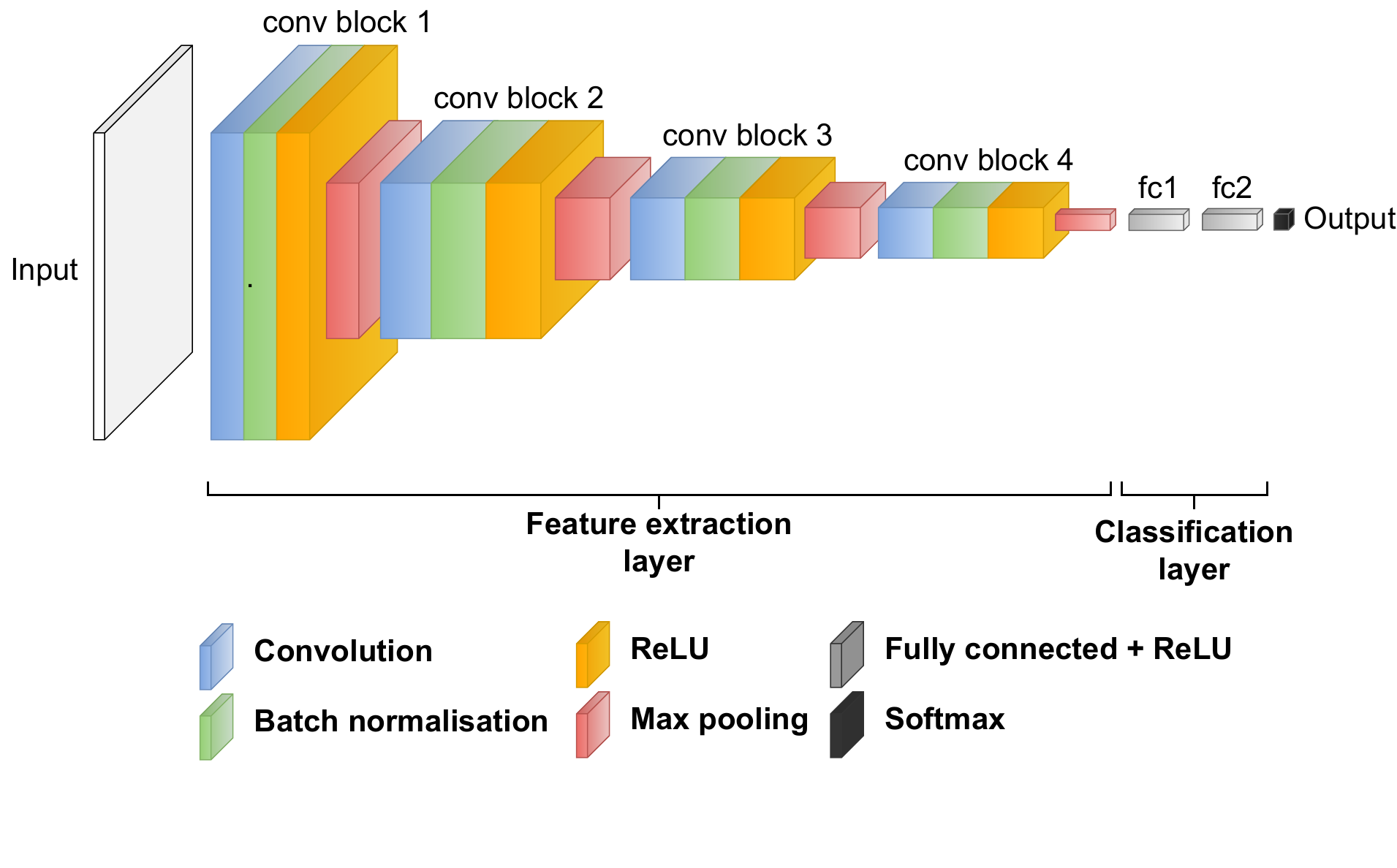}
\caption{The CloudScout~\cite{giuffrida2020cloudscout} architecture.}
\label{fig:cloudscout}
\end{figure}

One real-world example of a satellite-borne CNN-based cloud detector is CloudScout~\cite{giuffrida2020cloudscout}. As shown in Fig.~\ref{fig:cloudscout}, the architecture is made up of two core layers: feature extraction and classification. The feature extraction layer is made up of 4 blocks of convolutional layers, each having different number of filters, kernel sizes, batch normalisation and pooling operators, and ReLU activations. Whereas, the classification layer is made up of two fully connected layers with ReLU activations. The CNN takes as inputs, 3 bands of the preprocessed data cube and outputs a binary response of whether the data cube is \emph{cloudy} or \emph{not cloudy}.

% mention the use of models with more than 3 input bands and larger memory footprint
The use of 3 bands to perform cloud detection was simply due to the limitations of the compiler for the EoT AI board (i.e., it only supported inputs with a maximum of 3 bands). However, in Sec.~\ref{sec:results:domain_gap}, we will investigate the effects of domain gap by increasing the number of bands as well as using a more sophisticated CNN architecture (i.e., ResNet50~\cite{he2016deep}). Details of the cloud detectors that we investigated are provided in \autoref{tab:cloud_detectors}.

\begin{table}[ht]
    \small
    \centering
    \caption{Comparing cloud detectors in terms of (i) memory footprint (in MB using FP32 weights), (ii) number of input bands, (iii) number of weights in the convolutional (CONV) layer, (iv) number of weights in the batch normalisation (BN) layer, and (v) number of weights in the fully connected (FC) layer.}
    \label{tab:cloud_detectors}
    \ifarxiv
        \begin{tabularx}{1.02\linewidth}{p{1.76cm}>{\raggedleft\arraybackslash}p{0.71cm}>{\raggedleft\arraybackslash}p{0.85cm}>{\raggedleft\arraybackslash}p{1.42cm}>{\raggedleft\arraybackslash}p{0.9cm}>{\raggedleft\arraybackslash}p{1cm}R}
        \hline
        \makecell[c]{\textbf{Cloud}\\\textbf{detectors}} & \makecell[c]{\textbf{Mem}} & \makecell[c]{\textbf{No. of}\\\textbf{bands}} & \makecell[c]{\textbf{CONV}} & \makecell[c]{\textbf{BN}} & \makecell[c]{\textbf{FC}} & \makecell[c]{\textbf{Total}}  \\
        \hline
        CloudScout-3 & 5.2  & 3 & 1,026,560  & 2,304  & 263,682 & 1,292,546 \\
        CloudScout-8 & 5.2  & 8 & 1,042,560  & 2,304  & 263,682 & 1,308,546 \\
        ResNet50-3   & 94.0 & 3 & 23,454,912 & 53,120 & 4,098   & 23,512,130 \\
        ResNet50-8   & 94.0 & 8 & 23,470,592 & 53,120 & 4,098   & 23,527,810 \\
        \hline
        \end{tabularx}
    \else
        \begin{tabularx}{\linewidth}{lRRRRRR}
        \hline
        \textbf{Cloud detectors} & \makecell[c]{\textbf{Memory}\\\textbf{footprint}} & \makecell[c]{\textbf{No. of}\\\textbf{bands}} & \makecell[c]{\textbf{CONV}} & \makecell[c]{\textbf{BN}} & \makecell[c]{\textbf{FC}} & \makecell[c]{\textbf{Total}}  \\
        \hline
        CloudScout-3 & 5.20  & 3 & 1,026,560  & 2,304  & 263,682 & 1,292,546 \\
        CloudScout-8 & 5.20  & 8 & 1,042,560  & 2,304  & 263,682 & 1,308,546 \\
        ResNet50-3   & 94.00 & 3 & 23,454,912 & 53,120 & 4,098   & 23,512,130 \\
        ResNet50-8   & 94.00 & 8 & 23,470,592 & 53,120 & 4,098   & 23,527,810 \\
        \hline
        \end{tabularx}
    \fi
\end{table}

\subsection{Training cloud detectors}\label{sec:source_detector}\label{sec:build_det:train}  
Following \cite{giuffrida2020cloudscout}, a two-stage supervised training procedure was performed on the cloud detector $f(\cdot \, ; \, \theta)$ parameterised by $\theta = \{ \theta_{\text{ext}}, \theta_{\text{cls}}\}$, where  $\theta_{\text{ext}}$ denotes the weights in the feature extraction layer and $\theta_{\text{cls}}$ denotes the weights in the classification layer. Training commenced by optimising $\theta_{\text{ext}}$ on the training set of $\cD^s_\text{TH30}$ to allow the feature extraction layer to recognise ``cloud shapes'':
\begin{align}
   \theta_{\text{ext}}^s = \argmin_{\theta_{\text{ext}}} \sum_{(x_i^s, y_i^s) \in \cD^s_\text{TH30}} L \left( f(x_i^s \, ; \, \theta_\text{ext}, \theta_\text{cls}), y_i^s\right),
\end{align}
where $L$ is the binary cross-entropy loss function. Then, $\theta_{\text{cls}}$ was optimised on the training set of $\cD^s_\text{TH70}$ to fine-tune the classification layer while freezing the parameters in the feature extraction layer:
\begin{align}
    \theta_{\text{cls}}^s = \argmin_{\theta_{\text{cls}}} \sum_{(x_i^s, y_i^s) \in \cD^s_\text{TH70}} L \left( f(x_i^s \, ; \, \theta^s_\text{ext}, \theta_\text{cls}), y_i^s \right).
\end{align}
Other training specifications such as learning rate and its decay schedule, as well as loss function modifications followed \cite{giuffrida2020cloudscout}. Once trained, a source cloud detector $f(\cdot \, ; \, \theta^s)$ was obtained, where $\theta^s = \{\theta_{\text{ext}}^s, \theta_{\text{cls}}^s \}$. In Sec.~\ref{sec:model_performance}, we will describe how the performance of $f(\cdot \, ; \, \theta^s)$ was evaluated.

% The model performance of the cloud detectors in \autoref{tab:cloud_detectors} are provided in Sec.~\ref{sec:results:domain_gap}. 

% As discussed in Sec.~\ref{sec:pre-deployment}, $\theta^s$ is deployed onto the satellite and launched into orbit.

%%%%%%%%%%%%%%%%%%%%%%%%%%%%%%%%%%%%%%%%%%%%%%%%%%%%%%%%%%%%%%%%%%%%%%%%%%%%%%%%%%%%%%%%%%%%%%%%
\ifarxiv
    \vspace{-2ex}
\fi
\section{Adapting the cloud detector to the target domain}\label{sec:adapting}
In this section, we provide details of the steps involved in the post-deployment domain adaptation stage (Sec.~\ref{sec:post_deployment}). More specifically, we provide details of our proposed bandwidth-efficient SDA algorithm for offline adaptation (Sec.~\ref{sec:offline_sda}), and our solution for TTA on satellite hardware to achieve online adaptation (Sec.~\ref{sec:online_tta}). 

\subsection{Bandwidth-efficient SDA}\label{sec:offline_sda}
\label{sec:adapt_det:offline}
% In this adaptation setting, we assume there is network connectivity between the ground station and satellite and thus, adapting $f(\cdot \, ; \, \theta^s)$ to $\cD^t$ will be performed on the ground station by downlinking the target data captured by the satellite, labelling it, and then, updating $\theta_s$ on GPU using PyTorch. Finally, the model updates $\theta_t$ are up-linked when the satellite comes into view (data transfer window) with the ground station (see Fig.~\ref{fig:offline_adaptation} for an illustration). 

% Given $\cD^t = \{ x^t_j, y^t_j \}_{j=1}^{N^t}$, our $f(\cdot \, ; \, \theta^s)$ can be adapted as follows,
% \begin{align}
%    \theta^t = \argmin_{\theta^s} \sum_{(x_j, y_j) \in \cD^t} L \left( f(x_j \, ; \, \theta^s), y_j\right)
%    \label{eq:dense_update}
% \end{align}
% Eq.~\eqref{eq:dense_update} updates ...

% gradients represent the sensitivity of the model's output to changes in its parameters
To solve Problem 1 in Sec.~\ref{sec:offline_adaptation}, we employed the Fisher-Induced Sparse uncHanging (FISH) Mask~\cite{sung2021training} to select a small (or sparse) subset of $\theta^s$ denoted by $\hat{\theta}^s$  (i.e., $\hat{\theta}^s \subset \theta^s$) that are considered to be the ``most important" weights to update during the adaptation process. First, we measured the empirical Fisher information of $\theta^s$,
\begin{equation}\label{eq:fish}
    F_{\theta^s} = \frac{1}{\left|\cD^t_\text{TH70}\right|} \sum_{(x_j^t, y_j^t) \in \cD^t_\text{TH70}} \left( \nabla_{\theta^{s}} L\left( f^g(x_j^t \, ; \, \theta^s), y_j^t\right) \right)^2,
\end{equation}
where $\left|\cD^t_\text{TH70} \right|$ is the total number of training samples of $\cD^t_\text{TH70}$, and $\nabla$ is the gradient operator. Recall that $f^g(\cdot\,;\,\theta^s)$ is the ground copy of the source cloud detector. Eq.~\eqref{eq:fish} computes the vector $F_{\theta^s} \in \bbR^{\left| \theta^s\right|}$ and the importance of $\theta^s_k \in \theta^s$ is represented by a large value $F_{\theta_k^s}$. Then, given a desired mask sparsity level $l$, the subset $\hat{\theta}^s$ was obtained by selecting weights with the top $l$-highest Fisher values,
\begin{align}
    \hat{\theta}^s = \begin{Bmatrix}\, \theta^s_k \,\, \mid \,\, F_{\theta_k^s} \ge \texttt{sort}\left( F_{\theta^s}\right)_l \, \end{Bmatrix}.
\end{align}
Next, $\hat{\theta}^s$ was updated on $\cD^t$ as 
\begin{align}
   \hat{\theta}^t = \argmin_{\hat{\theta}^s} \sum_{(x_j^t, y_j^t) \in \cD^t_\text{TH70}} L \left( f^g(x_j^t \, ; \, \theta^s), y_j^t\right),
\end{align} 
while the remaining weights $\bar{\theta}^s = \theta^s \setminus \hat{\theta}^s$ were frozen. Lastly, $\theta_t$ was obtained by setting $\theta_t = \hat{\theta}^t \cup \bar{\theta}^s$. This algorithm allows us to uplink only $\hat{\theta}^t$ to update $f^d$. We will show in Sec.~\ref{sec:exp:fish} that updating only 25\% of the total weights of CloudScout, or 1\% of the total weights of ResNet50, is sufficient to achieve similar levels of performance as updating 100\% of the weights. 
    
% To save on memory usage and communication costs (i.e., uplink bandwidth), we employed the Fisher-induced sparse unchanging (FISH) mask~\cite{sung2021training} to select a subset of $\theta_s$ that are (in some sense) the most important to update during the adaptation process. The parameter's importance is measured by considering how much each parameter affects the model's output probabilities $\hat{y} = z(f_{\theta_s}(\cdot))$ where $z$ is the softmax operator. This can be measured by estimating the empirical Fisher information of each parameter:
% \begin{equation}  
%     \hat{F}(\hat{y}, \cD^t) =
%     \frac{1}{N} \sum_{i=1}^{N} \bigl( \nabla_{\theta_{s}} \log (\hat{y}) \bigl)^{2}
% \end{equation}
% where $\nabla$ is the gradient operator, and $N$ is the number of target samples used to calculate $\hat{F}$. If a given parameter heavily influences $\hat{y}$, then its corresponding entry in $\hat{F}$ will be large and vice versa. We then use this information to construct a FISH mask (i.e. binary mask where 1 means update the parameter and 0 means freeze the parameter) by simply selecting $k$ parameters with the largest Fisher information where $k$ is set according to the desired mask sparsity level. It is worth noting that computing the FISH mask is cheap since $\hat{F}$ can be computed efficiently using the backpropagation algorithm. Also, the mask only needs to be computed once since it is constructed before the adaptation process begins. Once the model has been adapted, $\theta_t$ and their respective indices are uplinked to the satellite.
 
\subsection{TTA on satellite hardware}\label{sec:online_tta}   
% Step 1 - build a runtime environment suitable for satellite compute hardware
% Andrew: Mention that XE1 can only run inference and thus unable to update models. Solution - A quad-core x86 64-bit CPU processor with 2 GB RAM was used to train/update Pytorch models on D-Orbit’s ION SCV004 satellite~\cite{ruzicka2023fast}

% \textcolor{red}{At this stage, we need to be specific with the hardware so definitely mention CogniSAT XE1 here}.\textcolor{green}{STATUS: Specifics of hardware added}

% To solve Problem 2 in Sec.~\ref{sec:online_adaptation}, we employ the Ubotica CogniSAT-XE1~\cite{2023ubotica} for on-board cloud detection. Ubotica CogniSAT-XE1 is a low-power edge processing device (i) featuring the Intel Myriad 2 VPU for accelerating machine-learning computation, including neural-network inferencing, and (ii) designed for SmallSat and CubeSat missions.

\begin{figure}[ht]
\centering
\includegraphics[width=\linewidth]{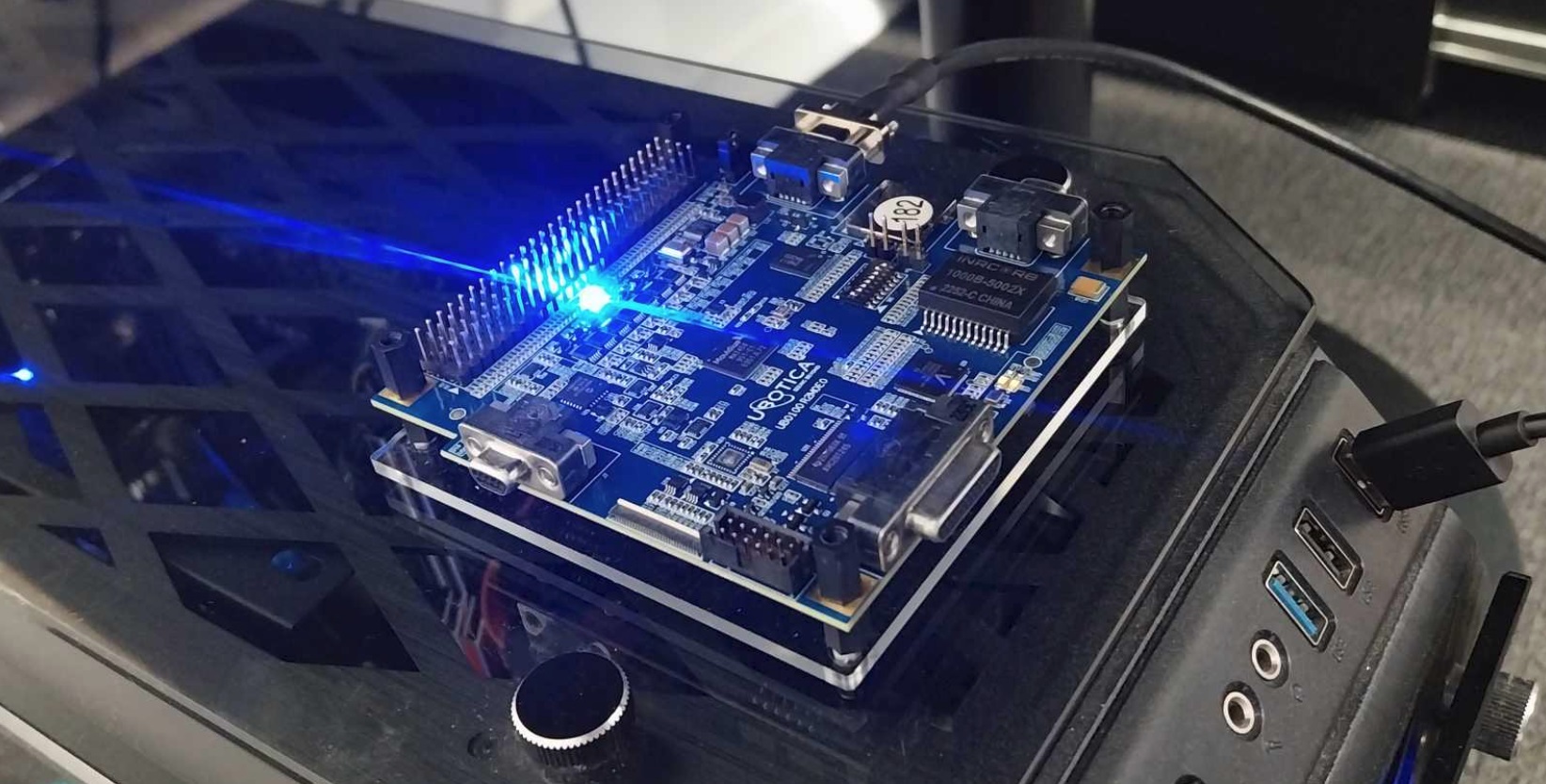}
\caption{Our Ubotica CognitSAT-XE1 connected to a standard Linux desktop.}
\label{fig:testbed}
\end{figure}

\begin{figure}[ht]
\centering
\includegraphics[width=\linewidth]{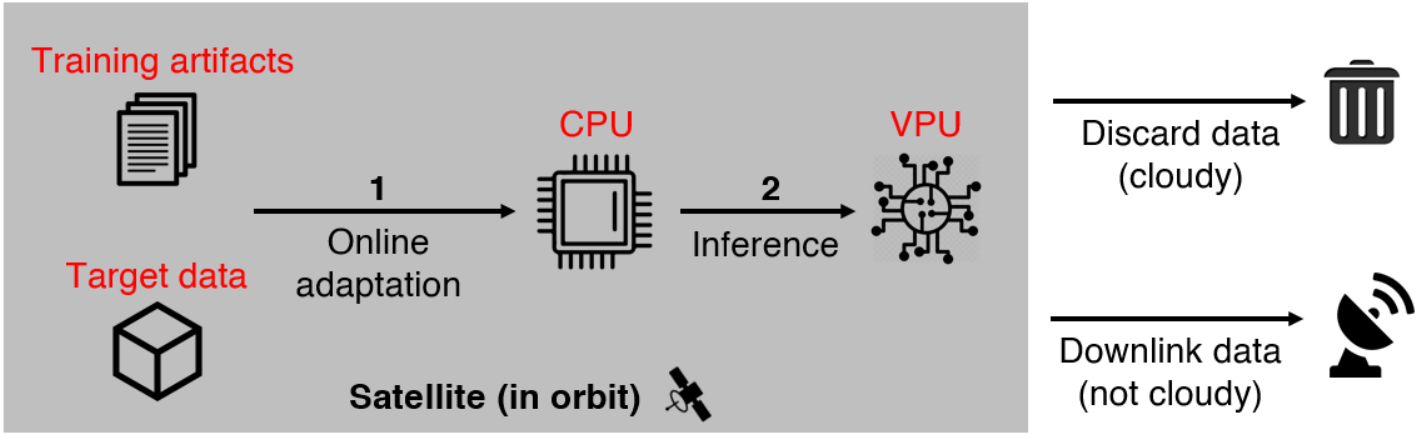}
\caption{Online adaptation and inference on the satellite in orbit.}
\label{fig:online_adaptation}
\end{figure}

To solve Problem 2 in Sec.~\ref{sec:online_adaptation}, we built and ran ONNX Runtime (ORT)~\cite{2023onnxruntime} on a standard Linux desktop with the Ubotica CognitSAT-XE1 connected via USB (see Fig.~\ref{fig:testbed}). Details of the XE1 will be provided in Sec.~\ref{sec:model_performance}. ORT was selected since it only requires $\approx$ 18.1 MB (version 1.15) of disk space and supports a wide range of operating systems and programming languages. Prior to deploying $f^d(\cdot \, ; \, \theta^s)$ onto the satellite in the pre-deployment model training phase (see Fig.~\ref{fig:offline_training}), the model was converted to the ONNX format, which was then used to generate training artefacts (i.e., training, evaluation and optimiser ONNX models, as well as checkpoint states). As shown in Fig.~\ref{fig:online_adaptation}, these training artefacts were then deployed onto the satellite and used to execute the TTA algorithms on CPU. 
% These training artefacts are then deployed onto the satellite in its pre-launch state and used to execute the TTA algorithms on CPU. 

% \textcolor{red}{TJ: A bit more background (a couple of sentences max) on the overall idea of TTA and Algorithm 1 is required. For example, tell the readers that the essence of how you do TTA is to update the batch norm layers (cite previous papers from ML that suggest this), and give reasonings why this is sensible. As it currently stands, you launch into ADAPT too abruptly. You are too focussed on detailing the mechanics, and ignored the intuitive ideas.}\textcolor{green}{STATUS: 3 sentences added}

One well-known TTA approach~\cite{wang2021tent, mirza2022norm, li2017revisiting, schneider2020improving} is to update the batch normalisation (BN) layers of a deep network (which in our case is the source cloud detector $f(\cdot \, ; \, \theta^s)$) in an unsupervised manner (using in our case the unlabelled target dataset $\tilde{\cD}^t$). The role of the BN layer is to normalise the intermediate outputs of each layer to zero mean and unit variance. However, this normalisation effect breaks when the source and target distributions significantly differ. As described in Algorithm~\ref{algo:online_adapt}, TTA is executed by the $\texttt{ADAPT}(\cdot)$ function but only when a batch of target samples $\cB$ is collected and reaches a certain (predefined) size $n_{\cB}$. This function is implemented by employing \acf{DUA}~\cite{mirza2022norm} (see Sec.~\ref{sec:online_adapt:dua}) and alternatively Test Entropy Minimisation (Tent)~\cite{wang2021tent} (see Sec.~\ref{sec:online_adapt:tent}) since both methods are efficient in terms of computing power (i.e., they do not rely on supervision or processing of source data) and memory usage (i.e., they do not rely on source data or a large batch of target data to be saved on the compute hardware of the satellite).

\begin{algorithm}
    \begin{algorithmic}
        \State $\theta^t \gets \theta^s$ \Comment{Copy source weights to target weights}
        \State $\cB \gets \emptyset$ \Comment{Initialise current batch}
        \For{$j \gets 1$ to $N^t$}
            \State $\cB \gets \cB \cup x_j^t$
            \If{$\left| \cB \right| = n_{\cB}$}
                \State $\theta^{t} \gets \texttt{ADAPT}(\theta^{t}, \cB)$
                \State $\cB \gets \emptyset$
            \EndIf
        \EndFor
    \end{algorithmic}
    \caption{TTA algorithm}
    \label{algo:online_adapt}
\end{algorithm}

\subsubsection{DUA}\label{sec:online_adapt:dua}
\ac{DUA}~\cite{mirza2022norm} updates the running means and running variances of the BN layers of $\theta^{t}$. More concretely, let us define $\hat{\mu}$, $\hat{\sigma}^2$, and $m$ as the running means, running variances, and momentum of an arbitrary BN layer of $\theta^{t}$ respectively. Furthermore, let $\mu$ and $\sigma^2$ be the mean and variance of the batch $\cB$. \ac{DUA} first updates the momentum of the BN layer,
\begin{align}
    m \gets m \cdot \omega + \delta,
\end{align}
where $\omega \in (0,1)$ is the predefined momentum decay parameter and $\delta$ defines the lower bound of the momentum. Then, the running mean $\hat{\mu}$ and running variance $\hat{\sigma}^2$ are updated as
\begin{align}
    \hat{\mu} &\gets (1-m) \cdot \hat{\mu} + m \cdot \mu, \\
    \hat{\sigma}^2 &\gets (1-m) \cdot \hat{\sigma}^2 + m \cdot \sigma^2.
\end{align}
The main idea of \ac{DUA} is to gradually decay the momentum $m$, because a fixed momentum can demonstrably destabilise or slow down the convergence of the adaptation process~\cite{mirza2022norm}.

\subsubsection{Tent}
\label{sec:online_adapt:tent}

Similar to DUA~\cite{mirza2022norm}, Tent~\cite{wang2021tent} also updates the the BN layers of $\theta^t$ but with one minor difference --- the affine transformation parameters are also updated. Recall $\hat{\mu}$ and $\hat{\sigma}$ are the running mean and running variance of an arbitrary BN layer of $\theta^t$. We further denote $\gamma$ and $\beta$ as the affine transformation parameters of the BN layer. Given $\cB$, Tent~\cite{wang2021tent} first estimates the Shannon entropy loss:
\begin{align}
    H = -\sum_{x^t_j \in \cB} \sum_{c=1}^C \hat{y}^t_{j,c} \cdot \log \hat{y}^t_{j,c},
\end{align}
where $C$ is total number of classes, $\hat{y}^t_j$ is the prediction of $x^t_j$, and $\hat{y}^t_{j,c}$ is the predicted probability of $\hat{y}^t_j$ of class $c$. Then, the BN layer is updated in terms of
\begin{alignat}{2}
    \hat{\mu} &\gets \bbE_{\cB}[x^t], &\quad \hat{\sigma}^2 &\gets \bbE_{\cB}[(x^t - \mu)^2], \\
    \gamma &\gets \gamma + \frac{\partial H}{\partial \gamma}, &\quad \beta &\gets \beta + \frac{\partial H}{\partial \beta}.
\end{alignat}
% \textcolor{red}{This paragraph seems hanging. At least add a short sentence to close it off, else the reader may think we forgot to say something here.}\textcolor{green}{STATUS: Closing sentence added}
In contrast to \cite{mirza2022norm}, Tent~\cite{wang2021tent} uses a fixed momentum, and the normalisation statistics $\hat{\mu}_0$, $\hat{\sigma}_{0}^{2}$ are recalculated from scratch on target data.

\ifarxiv
    \vspace{-2ex}
\fi
\section{Executing and evaluating the cloud detector on the Ubotica CogniSAT-XE1}\label{sec:model_performance}

% {\color{blue}
% In principle, inference performance is to be evaluated on the Ubotica CogniSAT-XE1~\cite{2023ubotica}, but the device is only used to evaluate online adaptation. For evaluating offline adaptation, an RTX 2070 GPU is used instead because the inference results are similar yet the GPU is XXX orders of magnitude faster.

% In order to execute and evaluate the performance of a cloud detector $f(\cdot\,;\,\theta)$ on the Ubotica, the model format must be converted to the Ubotica Neural Network (UNN) format. This conversion involves 3 steps: 
% \begin{enumerate}
%     \item Export the model to the ONNX format.
%     \item Convert the generated ONNX files to the OpenVINO Intermediate Representation format.
%     \item Convert the OpenVINO Intermediate Representation to the UNN format.
% \end{enumerate}
% Note that Step 2 also quantises $\theta$ from FP32 to FP16, which is required to run inference on the Intel Myriad 2 VPU in the Ubotica. For the GPU, $\theta$ remains in FP32.
% }
The Ubotica CogniSAT-XE1~\cite{2023ubotica} is a low-power edge processing device designed for SmallSat and CubeSat missions. It features the Intel Myriad 2 VPU and its main purpose is to accelerate machine learning inference. In order to execute a cloud detector $f(\cdot\,;\,\theta)$ on this device, the model format must be converted to the Ubotica Neural Network (UNN) format. This conversion involves 3 steps: 
\begin{enumerate}
    \itemsep0.5em 
    \item Export the model to the ONNX format.
    \item Convert the generated ONNX files to the OpenVINO Intermediate Representation format.
    \item Convert the OpenVINO Intermediate Representation to the UNN format.
\end{enumerate}
Note that Step 2 also quantises $\theta$ from FP32 to FP16, which is required to run inference on the Intel Myriad 2 VPU. The performance of $f(\cdot\,;\,\theta)$ on a dataset $\cD = \{x_m, y_m\}_{m=1}^M$ is evaluated based on two metrics:
\begin{itemize}
    \item Accuracy (ACC) of $f(\cdot\,;\,\theta)$,
        \begin{equation}\label{eq:accuracy}
            \text{ACC} \triangleq 
            \frac{1}{M}
            \sum^{M}_{m=1} \bbI(\arg \max \hat{y}_{m} = y_{m}) \times 100\%,
        \end{equation}
        where $\bbI(\cdot)$ is the indicator function and $\hat{y}_m = f(x_m\,;\,\theta)$. The higher the test accuracy means the higher $f(\cdot\,;\,\theta)$ predicts the correct class label. This helps in increasing the quality of each prediction especially in challenging situations, e.g., clouds on ice, or clouds on salt-lake.
    \item False positives (FP) of $f(\cdot\,;\,\theta)$,
        \begin{equation}\label{eq:average_probability}
            \text{FP}\triangleq 
            \frac{1}{M}
            \sum^{M}_{m=1} \bbI(\arg \max \hat{y}_{m} = 1 \land y_{m} = 0) \times 100\%.
        \end{equation}
        The lower the FP means the less $f(\cdot\,;\,\theta)$ incorrectly predicts non-cloudy data cubes as cloudy, which helps avoid discarding clear sky data cubes.
\end{itemize}
The performance of $f(\cdot\,;\,\theta)$ on source and target datasets (Sec.~\ref{sec:groundtruths}) can be evaluated by substituting $\cD$ with $\cD^s_\text{TH70}$ and $\cD^t_\text{TH70}$ respectively. In principle, inference performance is to be evaluated on the XE1 since it is the target device of interest. However, we found based on preliminary testing that there were minute differences in performance when we evaluated $f(\cdot\,;\,\theta)$ with FP32 weights on a Linux desktop and FP16 weights on the XE1.
% \hl{Mention that you evaluated the cloud detectors with FP32 weights on a desktop computer first to establish domain gap (Sec 8.1) and test adaptation approaches (8.2). Then, you evaluated the cloud detectors with FP16 weights on the XE1 (Sec 8.3). Also mention that ideally, you should of tested everything on the XE1 since it is the target device. However, based on preliminary testing, you found that there were minute differences in performance when evaluating on the desktop and XE1.}

\ifarxiv
    \vspace{-2ex}
\fi
\section{Results}\label{sec:results}
In this section, we first empirically establish the presence of nontrivial domain gap in the hyperspectral cloud detection task (Sec.~\ref{sec:results:domain_gap}). Then, we evaluate the performance of the proposed algorithms for offline and online domain adaptation (Sec.~\ref{sec:results:ablation}). 

\subsection{Domain gap in hyperspectral data}\label{sec:results:domain_gap}
We trained the cloud detector in \autoref{tab:cloud_detectors} on the source dataset and evaluated their performance on the target dataset, \emph{without} domain adaptation. For cloud detectors trained on 3 bands, we selected (i) bands 1, 2, 8a of Sentinel-2, and (ii) bands 1, 2, 5 of Landsat 9. Otherwise, we used the 8 shared bands of Sentinel-2 and Landsat 9. The different combinations of model settings are indicated by the naming convention
\begin{center}
    ARCH-NUMBANDS-SOURCE
\end{center}
where
\begin{itemize}
    \item ARCH is either the architecture of CloudScout or ResNet50,
    \item NUMBANDS is either 3 or 8,
    \item SOURCE is either S2 for Sentinel-2 or L9 for Landsat 9.
    % \item DATA is one of the training datasets in Sec.~\ref{sec:dataset_construction}.
\end{itemize}
For example, CloudScout-8-L9 refers to training the CloudScout architecture with 8 bands on Landsat 9 defined as the source domain. We also indicate the different datasets used to evaluate the source cloud detectors by the following naming convention:
\begin{center}
    SAT-SET
\end{center}
where
\begin{itemize}
    \item SAT is either S2 for Sentinel-2 or L9 for Landsat 9, 
    \item SET is either TRAIN for training set or TEST for testing set.
\end{itemize}
For example, S2-TEST means the testing set consisting of Sentinel-2 data. Note that the training and testing sets have ground truth labels obtained by applying a 70\% threshold to the cloud masks. Major findings that can be observed from the results in
\autoref{tab:domain_gap_1} and \autoref{tab:domain_gap_2} are:
\begin{itemize}
    \item The domain gap is more prominent in cloud detectors trained on 8 bands as compared to their 3-band counterparts; observe the difference in performance between CloudScout-3-S2 (ACC/FP of 66.40\%/0.80\%) and CloudScout-8-S2 (ACC/FP of 52.00\%/48.00\%) evaluated on L9-TEST. 
    \item The domain gap appears to be smaller in ResNet50 than CloudScout; observe the difference in performance between CloudScout-3-S2 (ACC/FP of 66.40\%/0.80\%) and ResNet50-3-S2 (ACC/FP of 90.80\%/1.60\%) evaluated on L9-TEST.
    \item Overall, the effects of domain gap are significant, since it prevents machine learning models from performing as required by EO mission standards e.g., CloudScout~\cite{giuffrida2020cloudscout} requires a minimum ACC of 85\% and maximum FP of 1.2\%.
\end{itemize}
These results were performed on cloud detectors with FP32 weights on a Linux desktop since quantising the weights to FP16 was found to have negligible effects on performance when evaluated on the XE1. The results also confirm the necessity of domain adaptation.

\begin{table*}[ht]
    % \scriptsize
    \small
    \centering    
    \caption{Model performance of cloud detectors trained on Sentinel-2 and evaluated on Landsat 9, without domain adaptation. GAP is the absolute difference in performance between the testing sets of Sentinel-2 and Landsat 9. Text in red and green show negative and positive effects on performance respectively.}
    \label{tab:domain_gap_1}
    % \begin{tabular}{|p{2.3cm}|p{1.4cm}|p{1.1cm}|p{1.4cm}|p{1.1cm}|p{1.4cm}|p{1.1cm}|p{1.4cm}|p{1.1cm}|}
    \begin{tabularx}{\linewidth}{lRRRRRRRR}
        \hline
        % \rowcolor{black}
        & \multicolumn{2}{c} {\Gape{\makecell[t]{\textcolor{black}{\textbf{S2-TRAIN}}}}} & \multicolumn{2}{c}{\Gape{\makecell[t]{\textcolor{black}{\textbf{S2-TEST}}}}} & \multicolumn{2}{c}{\Gape{\makecell[t]{\textcolor{black}{\textbf{L9-TEST}}}}} & \multicolumn{2}{c}{\Gape{\makecell[t]{\textcolor{black}{\textbf{GAP}}}}} \\
        % \hline
        \textbf{Model settings} & \makecell{\textbf{ACC} \\ \textbf{(\%)}} & \makecell{\textbf{FP} \\ \textbf{(\%)}} & \makecell{\textbf{ACC} \\ \textbf{(\%)}} & \makecell{\textbf{FP} \\ \textbf{(\%)}} & \makecell{\textbf{ACC} \\ \textbf{(\%)}} & \makecell{\textbf{FP} \\ \textbf{(\%)}} & \makecell{\textbf{ACC} \\ \textbf{(\%)}} & \makecell{\textbf{FP} \\ \textbf{(\%)}} \\
        \hline
        CloudScout-3-S2 & 92.85 & 0.96 & 92.07 & 1.72 & 66.40 & 0.80  & \textcolor{red}{25.67} & \textcolor{green}{0.92} \\
        CloudScout-8-S2 & 93.36 & 3.69 & 92.41 & 4.48 & 52.00 & 48.00 & \textcolor{red}{40.41} & \textcolor{red}{43.52} \\
        ResNet50-3-S2   & 97.86 & 1.11 & 93.10 & 4.14 & 90.80 & 1.60  & \textcolor{red}{2.30}  & \textcolor{green}{2.54} \\
        ResNet50-8-S2   & 93.73 & 2.51 & 93.79 & 2.41 & 56.80 & 43.20 & \textcolor{red}{36.99} & \textcolor{red}{40.79} \\
        \hline
    \end{tabularx}
\end{table*}

\begin{table*}[ht]
    % \scriptsize
    \small
    \centering    
    \caption{Model performance of cloud detectors trained on Landsat 9 and evaluated on Sentinel-2, without domain adaptation. GAP is the absolute difference in performance between the testing sets of Landsat 9 and Sentinel-2. Text in red and green show negative and positive effects on performance respectively.}
    \label{tab:domain_gap_2}
    % \begin{tabular}{|p{2.3cm}|p{1.4cm}|p{1.1cm}|p{1.4cm}|p{1.1cm}|p{1.4cm}|p{1.1cm}|p{1.4cm}|p{1.1cm}|}
    \begin{tabularx}{\linewidth}{lRRRRRRRR}
        \hline
        % \rowcolor{black}
        & \multicolumn{2}{c} {\Gape{\makecell[t]{\textcolor{black}{\textbf{L9-TRAIN}}}}} & \multicolumn{2}{c}{\Gape{\makecell[t]{\textcolor{black}{\textbf{L9-TEST}}}}} & \multicolumn{2}{c}{\Gape{\makecell[t]{\textcolor{black}{\textbf{S2-TEST}}}}} & \multicolumn{2}{c}{\Gape{\makecell[t]{\textcolor{black}{\textbf{GAP}}}}} \\
        % & \multicolumn{2}{|l|} {\textcolor{white}{\textbf{L9-TRAIN}}} & \multicolumn{2}{l|}{\textcolor{white}{\textbf{L9-TEST}}} & \multicolumn{2}{l|}{\textcolor{white}{\textbf{S2-TEST}}} & \multicolumn{2}{l|}{\textcolor{white}{\textbf{GAP}}} \\
        % \hline
        \textbf{Model settings} & \makecell{\textbf{ACC} \\ \textbf{(\%)}} & \makecell{\textbf{FP} \\ \textbf{(\%)}} & \makecell{\textbf{ACC} \\ \textbf{(\%)}} & \makecell{\textbf{FP} \\ \textbf{(\%)}} & \makecell{\textbf{ACC} \\ \textbf{(\%)}} & \makecell{\textbf{FP} \\ \textbf{(\%)}} & \makecell{\textbf{ACC} \\ \textbf{(\%)}} & \makecell{\textbf{FP} \\ \textbf{(\%)}} \\
        \hline
        CloudScout-3-L9 & 88.49 & 0.52 & 85.60 & 2.00 & 77.24 & 8.62  & \textcolor{red}{8.36}  & \textcolor{red}{6.62} \\
        CloudScout-8-L9 & 92.61 & 3.78 & 88.80 & 4.40 & 67.24 & 30.69 & \textcolor{red}{21.56} & \textcolor{red}{26.29} \\
        ResNet50-3-L9   & 95.10 & 2.66 & 90.80 & 4.40 & 83.79 & 12.07 & \textcolor{red}{7.01}  & \textcolor{red}{7.67} \\
        ResNet50-8-L9   & 98.20 & 1.80 & 93.60 & 4.80 & 61.03 & 37.93 & \textcolor{red}{32.57} & \textcolor{red}{33.13} \\
        \hline
    \end{tabularx}
\end{table*}

%%%%%%%%%%%%%%%%%%%%%%%%%%%%%%%%%%%%%%%%%%%%%%%%%%%%%%%%%%%%%%%%%%%%%%%%%%%%%%%%%%%%%%%%%%%%%%%%
\subsection{Ablation studies}\label{sec:results:ablation}
\label{sec:exp:ablation}
We present ablation studies in order to examine the approaches we employed for offline and online adaptation more carefully. These studies were also performed on cloud detectors with FP32 weights on a Linux desktop.

\subsubsection{Bandwidth-efficient SDA}\label{sec:exp:fish}
\begin{figure}[H]
    \centering
    \ifarxiv
        \includegraphics[width=\linewidth]{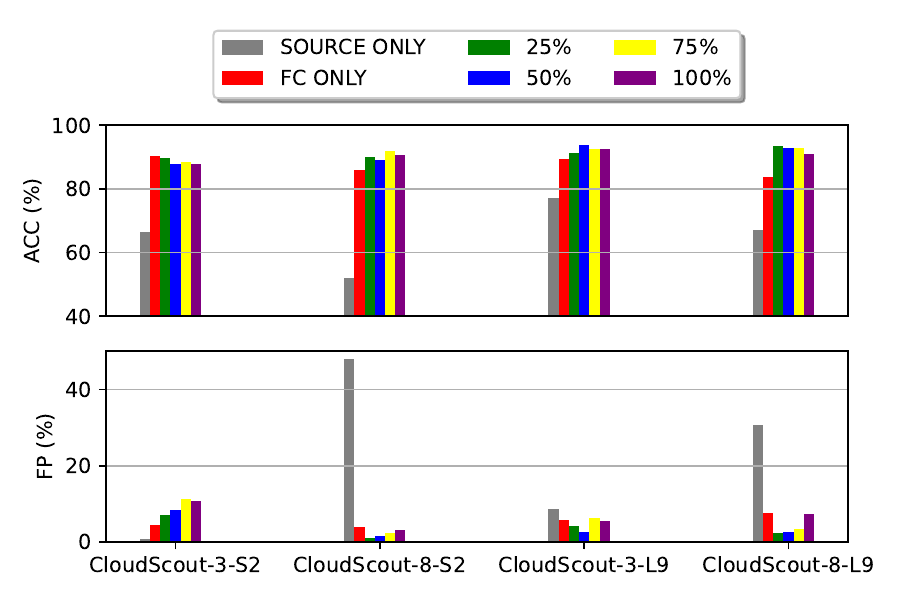}
        \vspace{-5ex}
    \else
        \includegraphics[width=0.90\linewidth]{./figures/fish_cloudscout.pdf}
    \fi
    \caption{Effects of FISH Mask on CloudScout models for different mask sparsity levels. SOURCE ONLY refers to Cloudscout models trained only on source data and FC ONLY means that only the weights in the FC layers were updated.}
    \label{fig:fish_cloudscout}
\end{figure}

\begin{figure}[H]
    \centering
    \ifarxiv
        \vspace{-2ex}
        \includegraphics[width=\linewidth]{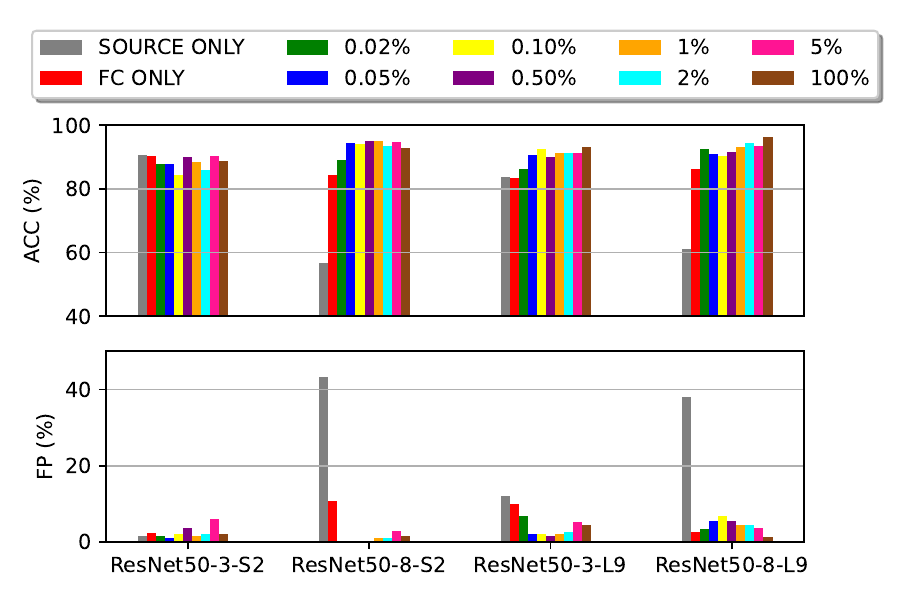}
        \vspace{-5ex}
    \else
        \includegraphics[width=0.90\linewidth]{./figures/fish_resnet50.pdf}
    \fi
    \caption{Effects of FISH Mask on ResNet50 models for different mask sparsity levels. SOURCE ONLY refers to ResNet50 models trained only on source data and FC ONLY means that only the weights in the FC layers were updated.}
    \label{fig:fish_resnet}
\end{figure}

The FISH Mask~\cite{sung2021training} was applied on the source cloud detectors (Sec.~\ref{sec:results:domain_gap}) with varying mask sparsity levels. The entire training set of the target dataset was used to estimate the Fisher information of each weight. We found that only a small subset of weights are required to achieve similar model performance as updating 100\% of the weights. Fig.~\ref{fig:fish_cloudscout} shows that for CloudScout models, only 25\% of the total weights needed to be updated, whereas Fig.~\ref{fig:fish_resnet} shows that for ResNet50 models, only 1\% were needed. The results indicate the usefulness of the FISH Mask to alleviate the uplink restrictions on model size, although this applies only to offline adaptation (Sec.~\ref{sec:offline_adaptation}) which assumes a labelled target dataset.

% \begin{figure}
%     \centering
%     \begin{minipage}{0.95\textwidth}
%         \centering
%         \includegraphics[width=0.90\linewidth]{./figures/fish_cloudscout.pdf}
%         \caption{Effects of FISH Mask on CloudScout models for different mask sparsity levels. SOURCE ONLY refers to Cloudscout models trained only on source data and FC ONLY means that only the weights in the FC layers were updated.}
%         \label{fig:fish_cloudscout}
%     \end{minipage}
%     \begin{minipage}{0.95\textwidth}
%         \centering
%         \includegraphics[width=0.90\linewidth]{./figures/fish_resnet50.pdf}
%         \caption{Effects of FISH Mask on ResNet50 models for different mask sparsity levels. SOURCE ONLY refers to ResNet50 models trained only on source data and FC ONLY means that only the weights in the FC layers were updated.}
%         \label{fig:fish_resnet}
%     \end{minipage}
% \end{figure}

%%%%%%%%%%%%%%%%%%%%%%%%%%%%%%%%%%%%%%%%%%%%%%%%%%%%%%%%%%%%%%%%%%%%%%%%%%%%%%%%%%%%%%%%%%%%%%%%
\subsubsection{TTA on satellite hardware}
First, DUA~\cite{mirza2022norm} was applied on CloudScout-3-S2 under different parameter settings, in particular, the number of samples used for adaptation and formation of small batches by performing data augmentations on each sample. 

% \begin{figure}
%     \centering
%     \begin{minipage}{0.95\textwidth}
%         \centering
%         \includegraphics[width=0.90\linewidth]{figures/dua_num_samples.pdf}
%         \caption{Effects of DUA on CloudScout-3-S2 with varying number of samples. SOURCE ONLY refers to CloudScout-3-S2 trained only on source data. Error bars represent standard deviations over 10 runs with different random seeds. No data augmentations were performed on each sample.}
%         \label{fig:dua_num_samples}
%     \end{minipage}
%     \begin{minipage}{0.95\textwidth}
%         \centering
%         \includegraphics[width=0.90\linewidth]{figures/dua_data_augmentations.pdf}
%         \caption{Effects of DUA on CloudScout-3-S2 with varying number of augmentation batch sizes. SOURCE ONLY refers to CloudScout-3-S2 trained only on source data. Error bars represent standard deviations over 10 runs with different random seeds. Number of samples was fixed to 16.}
%         \label{fig:dua_augmentations}
%     \end{minipage}
% \end{figure}  

\begin{figure}[H]
\centering
\ifarxiv
    \vspace{-2ex}
    \includegraphics[width=\linewidth]{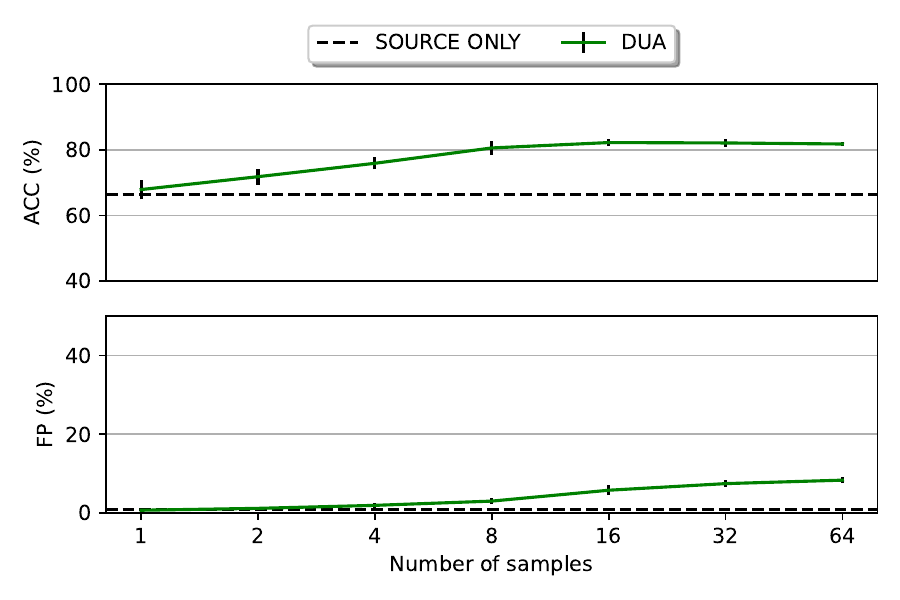}
    \vspace{-5ex}
\else
    \includegraphics[width=0.90\linewidth]{figures/dua_num_samples.pdf}
\fi
\caption{Effects of DUA on CloudScout-3-S2 with varying number of samples. SOURCE ONLY refers to CloudScout-3-S2 trained only on source data. Error bars represent standard deviations over 10 runs with different random seeds. No data augmentations were performed on each sample.}
\ifarxiv
    \vspace{-2ex}
\fi
\label{fig:dua_num_samples}
\end{figure}

\paragraph{Number of samples}
Fig.~\ref{fig:dua_num_samples} shows the effects of varying the number of samples used to update CloudScout-3-S2 to Landsat 9. We found that ACC saturates after forward passing 16 samples, however, at the expense of a higher FP rate. We also found that the ordering of the incoming samples holds little significance with very small deviations in model performance. More importantly, these results confirm that DUA is a memory-efficient approach since it only requires a small number of unlabelled target data to reach maximum performance. 

\begin{figure}[H]
\centering
\ifarxiv
    \vspace{-6ex}
    \includegraphics[width=\linewidth]{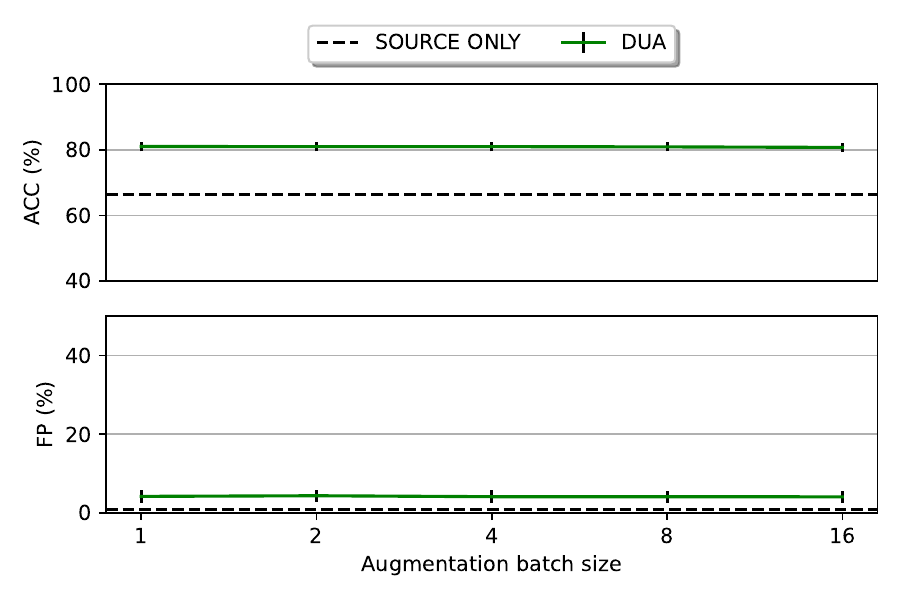}
    \vspace{-5ex}
\else
    \includegraphics[width=0.90\linewidth]{figures/dua_data_augmentations.pdf}
\fi
\caption{Effects of DUA on CloudScout-3-S2 with varying number of augmentation batch sizes. SOURCE ONLY refers to CloudScout-3-S2 trained only on source data. Error bars represent standard deviations over 10 runs with different random seeds. Number of samples was fixed to 16.}
\ifarxiv
    \vspace{-6ex}
\fi
\label{fig:dua_augmentations}
\end{figure}

\paragraph{Data augmentations}
Fig.~\ref{fig:dua_augmentations} shows the effects of varying the batch sizes that are formed by augmenting each sample with random horizontal flipping and rotations. Contrary to \cite{mirza2022norm}, we found that making small batches does not provide further improvements in model performance. This is good news because it eliminates the need to perform data augmentation and thus, reduces the time and computational efforts of DUA to perform adaptation.

\begin{figure}[H]
\centering
\ifarxiv
    \vspace{-6ex}
    \includegraphics[width=\linewidth]{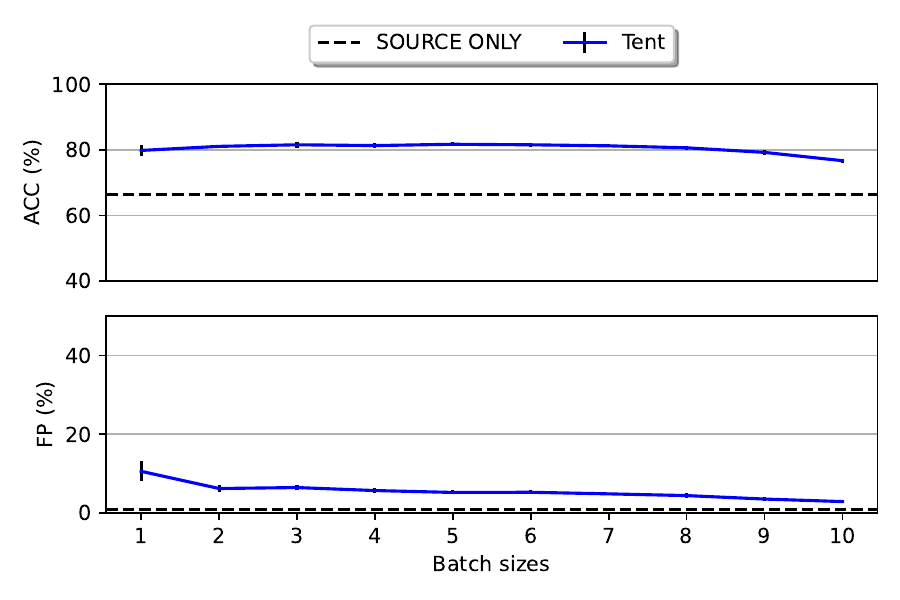}
    \vspace{-5ex}
\else
    \includegraphics[width=0.90\linewidth]{figures/tent_batch_sizes.pdf}
\fi
\caption{Effects of Tent on CloudScout-3-S2 with varying number of batch sizes. SOURCE ONLY refers to CloudScout-3-S2 trained only on source data. Error bars represent standard deviations over 10 runs with different random seeds. Number of epochs was fixed to 1.}
\ifarxiv
    \vspace{-8ex}
\fi
\label{fig:tent_batch_sizes}
\end{figure}

%%%%%%%%%%%%%%%%%%%%%%%%%%%%%%%%%%%%%%%%%%%%%%%%%%%%%%%%%%%%%%%%%%%%%%%%%%%%%%%%%%%%%%%%%%%%%%%%
Next, Tent~\cite{wang2021tent} was applied on CloudScout-3-S2 with varying batch sizes and number of epochs.
\ifarxiv
    \vspace{-12ex}
\fi
\paragraph{Batch size}
Fig.~\ref{fig:tent_batch_sizes} shows the effects of varying the batch sizes. We found that a batch size of 8 gives the best balance in performance in terms of increasing ACC and decreasing FP. Similar to DUA, Tent is also memory efficient since it does not need to wait for a large batch of target data.  

\begin{figure}[H]
\centering
\ifarxiv
    \vspace{-6ex}
    \includegraphics[width=\linewidth]{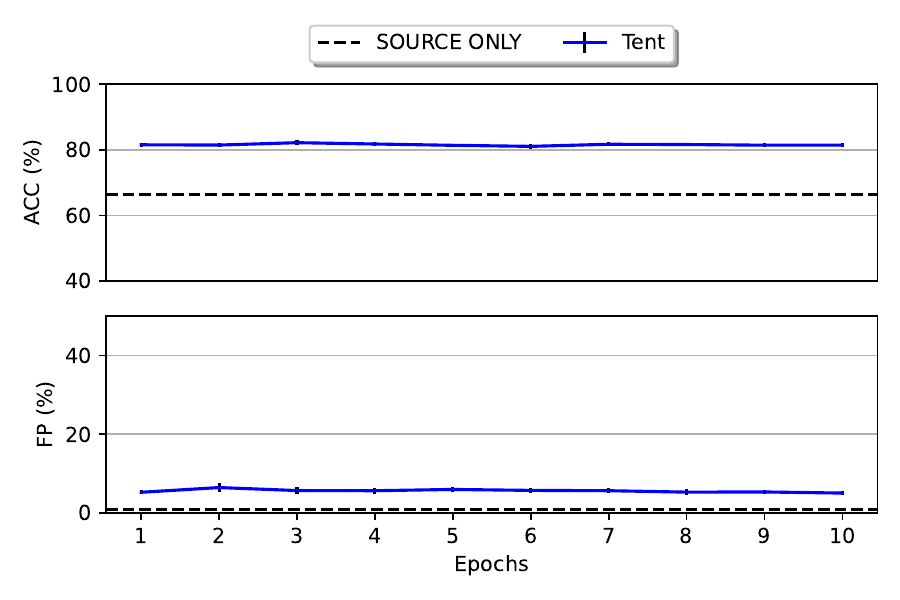}
    \vspace{-5ex}
\else
    \includegraphics[width=0.90\linewidth]{figures/tent_epochs.pdf}
\fi
\caption{Effects of Tent on CloudScout-3-S2 with varying number of epochs. SOURCE ONLY refers to CloudScout-3-S2 trained only on source data. Error bars represent standard deviations over 10 runs with different random seeds. Batch size was fixed to 8.}
\ifarxiv
    \vspace{-1ex}
\fi
\label{fig:tent_epochs}
\end{figure}

\paragraph{Number of epochs}
Fig.~\ref{fig:tent_epochs} shows no further improvements in performance beyond 1 epoch. This is good news because it eliminates the need to cycle through the same sample more than once and thus, reduces the time and computational efforts of Tent to perform adaptation to its maximum ability. 

% \begin{figure}
%     \centering
%     \begin{minipage}{0.95\textwidth}
%         \centering
%         \includegraphics[width=0.90\linewidth]{figures/tent_batch_sizes.pdf}
%         \caption{Effects of Tent on CloudScout-3-S2 with varying number of batch sizes. SOURCE ONLY refers to CloudScout-3-S2 trained only on source data. Error bars represent standard deviations over 10 runs with different random seeds. Number of epochs was fixed to 1.}
%         \label{fig:tent_batch_sizes}
%     \end{minipage}
%     \begin{minipage}{0.95\textwidth}
%         \centering
%         \includegraphics[width=0.90\linewidth]{figures/tent_epochs.pdf}
%         \caption{Effects of Tent on CloudScout-3-S2 with varying number of epochs. SOURCE ONLY refers to CloudScout-3-S2 trained only on source data. Error bars represent standard deviations over 10 runs with different random seeds. Batch size was fixed to 8.}
%         \label{fig:tent_epochs}
%     \end{minipage}
% \end{figure}

\begin{figure}[H]
\centering
\ifarxiv
    \vspace{-2ex}
    \includegraphics[width=\linewidth]{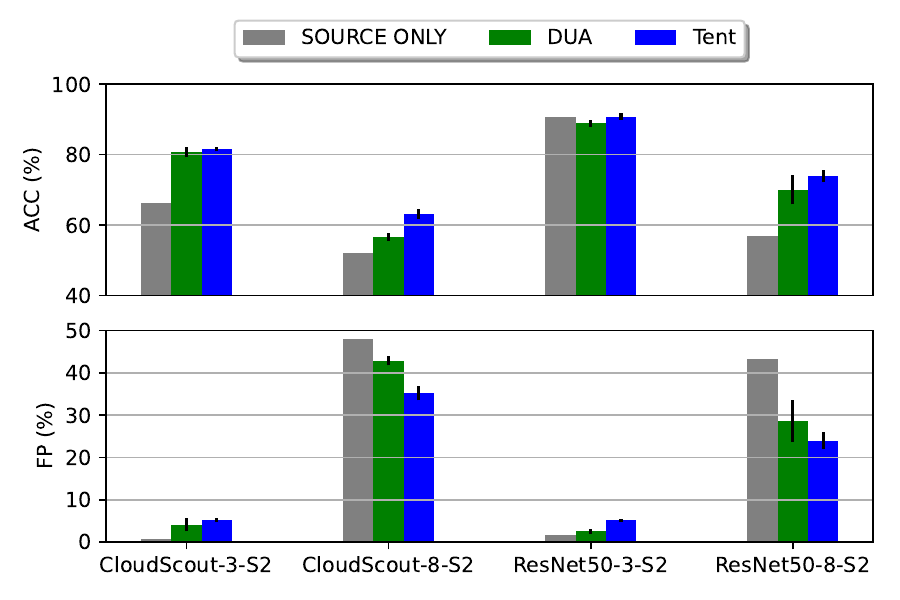}
    \vspace{-5ex}
\else
    \includegraphics[width=0.90\linewidth]{figures/dua_tent_s2_models.pdf}
\fi
\caption{Effects of DUA and Tent on the cloud detectors in \autoref{tab:domain_gap_1}. SOURCE ONLY refers to cloud detectors trained only on source data. Error bars represent standard deviations over 10 runs with different random seeds.}
\label{fig:s2_models_dua_tent}
\end{figure}

\begin{figure}[H]
\centering
\ifarxiv
    \vspace{-3ex}
    \includegraphics[width=\linewidth]{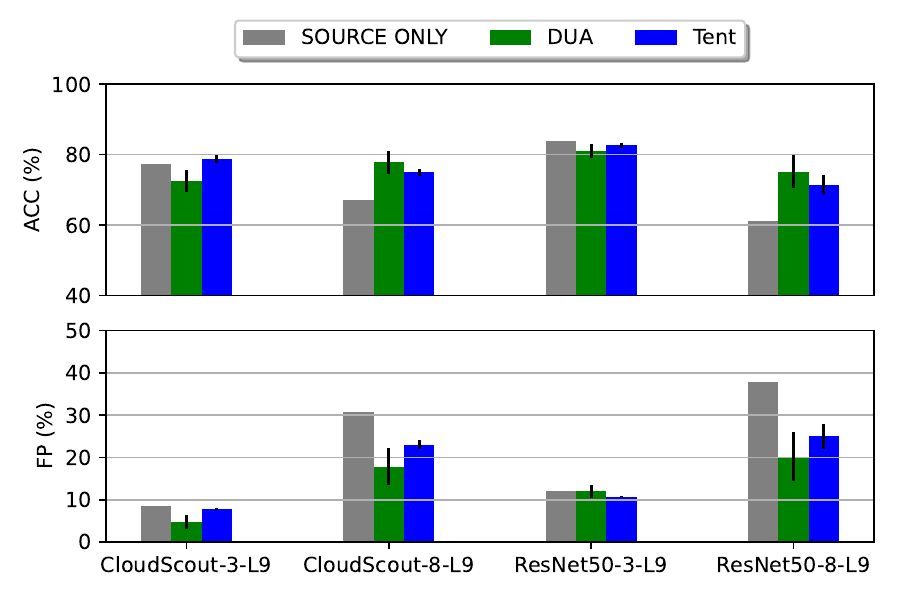}
    \vspace{-5ex}
\else
    \includegraphics[width=0.90\linewidth]{figures/dua_tent_l9_models.pdf}
\fi
\caption{Effects of DUA and Tent on the cloud detectors in \autoref{tab:domain_gap_2}. SOURCE ONLY refers to cloud detectors trained only on source data. Error bars represent standard deviations over 10 runs with different random seeds.}
\label{fig:l9_models_dua_tent}
\end{figure}

Lastly, we applied both TTA methods on the remaining source cloud detectors (Sec.~\ref{sec:results:domain_gap}). As shown in Fig.~\ref{fig:s2_models_dua_tent} and \ref{fig:l9_models_dua_tent}, Tent outperforms DUA in most cases but only by a slight margin. This was expected since DUA only updates the normalisation statistics in the BN layers. However, Tent requires more computational resources since it needs to perform backpropagation to update the affine transformation parameters. 

\begin{table*}[ht]
    \centering
    % \scriptsize
    \small
    \caption{Model performance of the source cloud detectors in \autoref{tab:domain_gap_1} with (FISH Mask, DUA and Tent) and without (SOURCE ONLY) domain adaptation on the Ubotica CogniSAT-XE1. Text in red and green show the negative and positive effects on performance respectively. The cloud detectors were also compared in terms of: (i) memory footprint (in MB using FP16 weights), and (ii) inference time (in ms per sample).}
    \label{tab:xe1_l9}
    % \begin{tabular}{|p{2.3cm}|r|r|r|r|r|r|r|r|r|r|}
    \begin{tabularx}{\linewidth}{lrRRRRRRRRR}
        \hline
        % \rowcolor{black}
        & & & \multicolumn{2}{c} {\Gape{\makecell[t]{\textcolor{black}{\textbf{SOURCE}} \\ \textcolor{black}{\textbf{ONLY}}}}} & \multicolumn{2}{c}{\textcolor{black}{\textbf{FISH Mask}}} & \multicolumn{2}{c}{\textcolor{black}{\textbf{DUA}}} & \multicolumn{2}{c}{\textcolor{black}{\textbf{Tent}}} \\
        % \hline
        \textbf{Model settings} & \makecell[t]{\textbf{Memory} \\ \textbf{footprint}} & \makecell{\textbf{Time}} & \makecell{\textbf{ACC} \\ \textbf{(\%)}} & \makecell{\textbf{FP} \\ \textbf{(\%)}} & \makecell{\textbf{ACC} \\ \textbf{(\%)}} & \makecell{\textbf{FP} \\ \textbf{(\%)}} & \makecell{\textbf{ACC} \\ \textbf{(\%)}} & \makecell{\textbf{FP} \\ \textbf{(\%)}} & \makecell{\textbf{ACC} \\ \textbf{(\%)}} & \makecell{\textbf{FP} \\ \textbf{(\%)}} \\
        \hline
        CloudScout-3-S2 & 2.60   & 2,252 & 
        66.40 & 0.80  & 
        \textcolor{green}{89.20} & \textcolor{red}{7.60} & 
        \textcolor{green}{79.20} & \textcolor{red}{3.60}  & 
        \textcolor{green}{81.20} & \textcolor{red}{5.20}  \\
        CloudScout-8-S2 & 2.60   & 2,015 & 
        52.00 & 48.00 & 
        \textcolor{green}{90.00} & \textcolor{green}{1.20} & 
        \textcolor{green}{57.20} & \textcolor{green}{42.40} & 
        \textcolor{green}{63.60} & \textcolor{green}{34.80} \\
        ResNet50-3-S2   & 47.00  & 1,245 & 
        90.80 & 1.60  & 
        \textcolor{red}{88.40} & 1.60 & 
        \textcolor{red}{88.00} & \textcolor{red}{2.80}  & 
        90.80 & \textcolor{red}{5.20}  \\
        ResNet50-8-S2 & 47.00  & 1,346 & 
        56.40 & 43.60 & 
        \textcolor{green}{95.20} & \textcolor{green}{1.20} & 
        \textcolor{green}{68.00} & \textcolor{green}{31.20} & 
        \textcolor{green}{70.80} & \textcolor{green}{27.60} \\
        \hline
    \end{tabularx}

    \vspace{0.3cm}
    \caption{Model performance of the source cloud detectors in \autoref{tab:domain_gap_2} with (FISH Mask, DUA and Tent) and without (SOURCE ONLY) domain adaptation, on the Ubotica CogniSAT-XE1. Text in red and green show the negative and positive effects on performance respectively. The cloud detectors were also compared in terms of: (i) memory footprint (in MB using FP16 weights), and (ii) inference time (in ms per sample).}
    \label{tab:xe1_s2}
    
    \begin{tabularx}{\linewidth}{lrRRRRRRRRR}
        \hline
        % \rowcolor{black}
        & & & \multicolumn{2}{c} {\Gape{\makecell[t]{\textcolor{black}{\textbf{SOURCE}} \\ \textcolor{black}{\textbf{ONLY}}}}} & \multicolumn{2}{c}{\textcolor{black}{\textbf{FISH Mask}}} & \multicolumn{2}{c}{\textcolor{black}{\textbf{DUA}}} & \multicolumn{2}{c}{\textcolor{black}{\textbf{Tent}}} \\
        % \hline
        \textbf{Model settings} & \makecell[t]{\textbf{Memory} \\ \textbf{footprint}} & \makecell{\textbf{Time}} & \makecell{\textbf{ACC} \\ \textbf{(\%)}} & \makecell{\textbf{FP} \\ \textbf{(\%)}} & \makecell{\textbf{ACC} \\ \textbf{(\%)}} & \makecell{\textbf{FP} \\ \textbf{(\%)}} & \makecell{\textbf{ACC} \\ \textbf{(\%)}} & \makecell{\textbf{FP} \\ \textbf{(\%)}} & \makecell{\textbf{ACC} \\ \textbf{(\%)}} & \makecell{\textbf{FP} \\ \textbf{(\%)}} \\
        \hline
        CloudScout-3-L9 & 2.60   & 2,252 & 
        77.20 & 8.40  & 
        \textcolor{green}{91.20} & \textcolor{green}{4.00} & 
        \textcolor{red}{72.00}   & \textcolor{red}{6.00}  & 
        \textcolor{red}{76.00}   & \textcolor{red}{7.60}  \\
        CloudScout-8-L9 & 2.60   & 2,015 & 
        68.80 & 28.80 & 
        \textcolor{green}{92.80}  & \textcolor{green}{2.40} & 
        \textcolor{green}{80.40}  & \textcolor{green}{14.00} & 
        \textcolor{green}{78.00}  & \textcolor{green}{19.20} \\
        ResNet50-3-L9   & 47.00   & 1,245 & 
        84.00 & 11.20 & 
        \textcolor{green}{91.20} & \textcolor{green}{2.00} & 
        \textcolor{red}{79.60}   & \textcolor{green}{10.00} & 
        \textcolor{red}{82.80}   & \textcolor{green}{10.00} \\
        ResNet50-8-L9   & 47.00  & 1,346 & 
        62.40 & 36.40 & 
        \textcolor{green}{94.00} & \textcolor{green}{3.20} & 
        \textcolor{green}{78.00} & \textcolor{green}{17.20} & 
        \textcolor{green}{77.60} & \textcolor{green}{18.00} \\
        \hline
    \end{tabularx}
\end{table*}

% \begin{figure}
%     \centering
%     \begin{minipage}{0.90\textwidth}
%         \centering
%         \includegraphics[width=\linewidth]{figures/dua_tent_s2_models.pdf}
%         \caption{Effects of DUA and Tent on the cloud detectors in \autoref{tab:domain_gap_1}. SOURCE ONLY refers to cloud detectors trained only on source data. Error bars represent standard deviations over 10 runs with different random seeds.}
%         \label{fig:s2_models_dua_tent}
%     \end{minipage}
%     \begin{minipage}{0.90\textwidth}
%         \includegraphics[width=\linewidth]{figures/dua_tent_l9_models.pdf}
%         \caption{Effects of DUA and Tent on the cloud detectors in \autoref{tab:domain_gap_2}. SOURCE ONLY refers to cloud detectors trained only on source data. Error bars represent standard deviations over 10 runs with different random seeds.}
%         \label{fig:l9_models_dua_tent}
%     \end{minipage}
% \end{figure}

\subsection{Performance on the Ubotica CogniSAT-XE1}\label{sec:exp:xe1}
Once the source and target cloud detectors have been obtained, we executed them on the Ubotica CogniSAT-XE1~\cite{2023ubotica} and evaluated their performance. In \autoref{tab:xe1_l9} and \autoref{tab:xe1_s2}, our results confirm that:
\begin{itemize}
    \item For offline adaptation, we can solve Problem 1 in Sec.~\ref{sec:offline_adaptation} by employing the FISH Mask and thereby, enabling more sophisticated models to be deployed and updated remotely through the thin uplink channel. As expected, we found that this adaptation approach outperforms the TTA approaches by a long margin. Note that CloudScout and ResNet50 models with a mask sparsity level of 25\% and 1\% respectively were only evaluated here. 
    \item For online adaptation, we can solve Problem 2 in Sec.~\ref{sec:online_adaptation} by employing DUA or Tent and thereby, establishing the viability of TTA on satellite-borne edge compute hardware. It is worth mentioning that even though these TTA approaches were able to reduce the domain gap, they may not be effective enough to warrant its use for onboard updating of models especially if there are certain performance requirements that need to be met. 
\end{itemize}

Other findings from our results are:
\begin{itemize}
    \item Quantising the weights from FP32 to FP16 had negligible effects on model performance as well as the added benefit of reducing the memory footprint by two-fold.
    \item ResNet50 models had a faster inference time (per sample) than CloudScout models which is surprising since they are $\approx$ 18x larger in size.   
\end{itemize}

\section{Conclusions and future work}

We showed the existence of domain gap when training a real-world CNN-based hyperspectral cloud detector on data from one EO mission and evaluating it on data from another mission. To address the domain gap, we proposed domain adaptation tasks framed in two different settings of an EO mission: (i) offline adaptation and (ii) online adaptation. For offline adaptation, our results show that only a small fraction of weights need to be updated (in a supervised manner) without noticeable impacts on performance. Offline adaptation enables more sophisticated and robust models to be deployed and remotely updated. Whereas for online adaptation, our results show the viability of test-time adaptation algorithms on space hardware. This enables us to directly update models onboard in an unsupervised manner. For future work, we plan on investigating other means of performing online adaptation that can satisfy the requirements in accordance to EO mission standards. 

\section*{Acknowledgements}

This work has been supported by the SmartSat CRC, whose activities are funded by the Australian Government’s CRC Program. Tat-Jun Chin is SmartSat CRC Professorial Chair of Sentient Satellites.

%%%%%%%%%%%%%%%%%%%%%%%%%%%%%%%%%%%%%%%%%%%%%%%%%%%%%%%%%%%%%%%%%%%%%%%%%%%%%%%%%%%%%%%%%%%%%%%%
%%%%%%%%%%%%%%%%%%%%%%%%%%%%%%%%%%%%%%%%%%%%%%%%%%%%%%%%%%%%%%%%%%%%%%%%%%%%%%%%%%%%%%%%%%%%%%%%
% \bibliographystyle{elsarticle-num} 
% \bibliography{IEEEabrv,references}

%% file: arxiv.bbl
\begin{thebibliography}{10}
\expandafter\ifx\csname url\endcsname\relax
  \def\url#1{\texttt{#1}}\fi
\expandafter\ifx\csname urlprefix\endcsname\relax\def\urlprefix{URL }\fi
\expandafter\ifx\csname href\endcsname\relax
  \def\href#1#2{#2} \def\path#1{#1}\fi

\bibitem{manfreda2018use}
S.~Manfreda, M.~F. McCabe, P.~E. Miller, R.~Lucas, V.~Pajuelo~Madrigal,
  G.~Mallinis, E.~Ben~Dor, D.~Helman, L.~Estes, G.~Ciraolo, J.~Müllerová,
  F.~Tauro, M.~I. De~Lima, J.~L. M.~P. De~Lima, A.~Maltese, F.~Frances,
  K.~Caylor, M.~Kohv, M.~Perks, G.~Ruiz-Pérez, Z.~Su, G.~Vico, B.~Toth, On the
  use of unmanned aerial systems for environmental monitoring, Remote Sensing
  10~(4) (2018).
\newblock \href {https://doi.org/10.3390/rs10040641}
  {\path{doi:10.3390/rs10040641}}.

\bibitem{2021sentinel-2}
{ESA},
  \href{https://sentinel.esa.int/web/sentinel/missions/sentinel-2}{Sentinel-2
  mission guide}, Sentinel Online, accessed 5 Sep 2021 (2021).
\newline\urlprefix\url{https://sentinel.esa.int/web/sentinel/missions/sentinel-2}

\bibitem{2023landsat9}
{NASA}, \href{https://landsat.gsfc.nasa.gov/satellites/landsat-9/}{Landsat 9},
  Landsat Science, accessed 3 Apr 2023 (2023).
\newline\urlprefix\url{https://landsat.gsfc.nasa.gov/satellites/landsat-9/}

\bibitem{esa-phisat-1}
{European Space Agency},
  \href{https://directory.eoportal.org/web/eoportal/satellite-missions/p/phisat-1}{{PhiSat-1
  Nanosatellite Mission}}, Satellite Missions Catalogue, eoPortal (Jun. 2020).
\newline\urlprefix\url{https://directory.eoportal.org/web/eoportal/satellite-missions/p/phisat-1}

\bibitem{esposito2019in-orbit}
M.~Esposito, S.~S. Conticello, M.~Pastena, B.~C. Domínguez, In-orbit
  demonstration of artificial intelligence applied to hyperspectral and thermal
  sensing from space, in: CubeSats and SmallSats for Remote Sensing III, 2019.

\bibitem{deniz2017eyes}
O.~Deniz, N.~Vallez, J.~L. Espinosa-Aranda, J.~M. Rico-Saavedra,
  J.~Parra-Patino, G.~Bueno, D.~Moloney, A.~Dehghani, A.~Dunne, A.~Pagani,
  S.~Krauss, R.~Reiser, M.~Waeny, M.~Sorci, T.~Llewellynn, C.~Fedorczak,
  T.~Larmoire, M.~Herbst, A.~Seirafi, K.~Seirafi, Eyes of things, Sensors
  17~(5) (2017).
\newblock \href {https://doi.org/10.3390/s17051173}
  {\path{doi:10.3390/s17051173}}.

\bibitem{giuffrida2020cloudscout}
G.~Giuffrida, L.~Diana, F.~de~Gioia, G.~Benelli, G.~Meoni, M.~Donati,
  L.~Fanucci, {CloudScout}: A deep neural network for on-board cloud detection
  on hyperspectral images, Remote Sensing 12~(14) (2020).
\newblock \href {https://doi.org/10.3390/rs12142205}
  {\path{doi:10.3390/rs12142205}}.

\bibitem{giuffrida2022phisat1}
G.~Giuffrida, L.~Fanucci, G.~Meoni, M.~Batič, L.~Buckley, A.~Dunne, C.~van
  Dijk, M.~Esposito, J.~Hefele, N.~Vercruyssen, G.~Furano, M.~Pastena,
  J.~Aschbacher, The $\phi$-sat-1 mission: The first on-board deep neural
  network demonstrator for satellite earth observation, {IEEE} Trans. Geosci.
  Remote Sens. 60 (2022) 1--14.
\newblock \href {https://doi.org/10.1109/TGRS.2021.3125567}
  {\path{doi:10.1109/TGRS.2021.3125567}}.

\bibitem{kouw2019introduction}
W.~M. Kouw, M.~Loog, An introduction to domain adaptation and transfer
  learning, arXiv preprint arXiv:1812.11806 (2019).
\newblock \href {https://doi.org/10.48550/arXiv.1812.11806}
  {\path{doi:10.48550/arXiv.1812.11806}}.

\bibitem{2021S2userhandbook}
{ESA}, Sentinel-2 user handbook,
  \small\url{https://sentinel.esa.int/documents/247904/685211/Sentinel-2_User_Handbook},
  accessed 3 Apr 2023 (2023).

\bibitem{2023L9datausershandbook}
{NASA}, Landsat 9 data users handbook,
  \small\url{https://d9-wret.s3.us-west-2.amazonaws.com/assets/palladium/production/s3fs-public/media/files/LSDS-2082_L9-Data-Users-Handbook_v1.pdf},
  accessed 3 Apr 2023 (2023).

\bibitem{2023hyperscout-2}
{Cosine}, Hyperscout 2, \small\url{https://www.cosine.nl/cases/hyperscout-2/},
  accessed 3 Apr 2023 (2023).

\bibitem{levinson2013automatic}
J.~Levinson, S.~Thrun,
  \href{https://www.roboticsproceedings.org/rss09/p29.pdf}{Automatic online
  calibration of cameras and lasers}, in: Robotics: Science and Systems, 2013.
\newline\urlprefix\url{https://www.roboticsproceedings.org/rss09/p29.pdf}

\bibitem{Ma2020ameliorating}
D.~Ma, Ameliorating environmental effects on hyperspectral images for improved
  phenotyping in greenhouse and field conditions, Ph.D. thesis, Agricultural
  and Biological Engineering, Purdue University (2020).

\bibitem{ganin2016domain}
Y.~Ganin, E.~Ustinova, H.~Ajakan, P.~Germain, H.~Larochelle, F.~Laviolette,
  M.~March, V.~Lempitsky, Domain-adversarial training of neural networks,
  Journal of Machine Learning Research 17~(59) (2016) 1--35.

\bibitem{2023ubotica}
{Ubotica},
  \href{https://ubotica.com/product/cognisat-xe1-product-overview/}{{Ubotica
  CogniSAT-XE1}}, accessed 7 Feb 2023 (2023).
\newline\urlprefix\url{https://ubotica.com/product/cognisat-xe1-product-overview/}

\bibitem{madry2017electrooptical}
S.~Madry, J.~N. Pelton, Electro-optical and hyperspectral remote sensing, in:
  Handbook of Satellite Applications, 2nd Edition, Springer International
  Publishing, 2017.

\bibitem{transon2018survey}
J.~Transon, R.~D’Andrimont, A.~Maugnard, P.~Defourny, Survey of hyperspectral
  earth observation applications from space in the {Sentinel-2} context, Remote
  Sensing 10~(2) (2018).
\newblock \href {https://doi.org/10.3390/rs10020157}
  {\path{doi:10.3390/rs10020157}}.

\bibitem{jeppesen2019cloud}
J.~H. Jeppesen, R.~H. Jacobsen, F.~Inceoglu, T.~S. Toftegaard, A cloud
  detection algorithm for satellite imagery based on deep learning, Remote
  Sensing of Environment 229 (2019) 247--259.
\newblock \href {https://doi.org/10.1016/j.rse.2019.03.039}
  {\path{doi:10.1016/j.rse.2019.03.039}}.

\bibitem{li2018onboard}
H.~Li, H.~Zheng, C.~Han, H.~Wang, M.~Miao, Onboard spectral and spatial cloud
  detection for hyperspectral remote sensing images, Remote Sensing 10~(1)
  (2018).
\newblock \href {https://doi.org/10.3390/rs10010152}
  {\path{doi:10.3390/rs10010152}}.

\bibitem{li2019cloud}
X.~Li, L.~Wang, Q.~Cheng, P.~Wu, W.~Gan, L.~Fang, Cloud removal in remote
  sensing images using nonnegative matrix factorization and error correction,
  ISPRS Journal of Photogrammetry and Remote Sensing 148 (2019) 103--113.
\newblock \href {https://doi.org/10.1016/j.isprsjprs.2018.12.013}
  {\path{doi:10.1016/j.isprsjprs.2018.12.013}}.

\bibitem{meraner2020cloud}
A.~Meraner, P.~Ebel, X.~X. Zhu, M.~Schmitt, {Cloud removal in Sentinel-2
  imagery using a deep residual neural network and SAR-optical data fusion},
  ISPRS Journal of Photogrammetry and Remote Sensing 166 (2020) 333--346.
\newblock \href {https://doi.org/10.1016/j.isprsjprs.2020.05.013}
  {\path{doi:10.1016/j.isprsjprs.2020.05.013}}.

\bibitem{zi2021thin}
Y.~Zi, F.~Xie, N.~Zhang, Z.~Jiang, W.~Zhu, H.~Zhang, Thin cloud removal for
  multispectral remote sensing images using convolutional neural networks
  combined with an imaging model, {IEEE} J. Sel. Topics Appl. Earth Observ.
  Remote Sens. 14 (2021) 3811--3823.
\newblock \href {https://doi.org/10.1109/JSTARS.2021.3068166}
  {\path{doi:10.1109/JSTARS.2021.3068166}}.

\bibitem{sinergise-cloud-masks}
{Sinergise Laboratory},
  \href{https://docs.sentinel-hub.com/api/latest/user-guides/cloud-masks/}{Cloud
  masks}, Sentinel Hub User Guide, accessed 26 Oct 2021 (2021).
\newline\urlprefix\url{https://docs.sentinel-hub.com/api/latest/user-guides/cloud-masks/}

\bibitem{lopezpuigdollers2021benchmarking}
D.~L\'{o}pez-Puigdollers, G.~Mateo-Garc\'{i}a, L.~G\'{o}mez-Chova, Benchmarking
  deep learning models for cloud detection in {Landsat-8} and {Sentinel-2}
  images, Remote Sensing 13~(5) (2021).
\newblock \href {https://doi.org/10.3390/rs13050992}
  {\path{doi:10.3390/rs13050992}}.

\bibitem{liu2021dcnet}
Y.~Liu, W.~Wang, Q.~Li, M.~Min, Z.~Yao, {DCNet}: A deformable convolutional
  cloud detection network for remote sensing imagery, {IEEE} Geosci. Remote
  Sens. Lett. (2021) 1--5\href {https://doi.org/10.1109/LGRS.2021.3086584}
  {\path{doi:10.1109/LGRS.2021.3086584}}.

\bibitem{li2019deep-ieee}
S.~Li, W.~Song, L.~Fang, Y.~Chen, P.~Ghamisi, J.~A. Benediktsson, Deep learning
  for hyperspectral image classification: An overview, {IEEE} Trans. Geosci.
  Remote Sens. 57~(9) (2019) 6690--6709.

\bibitem{long2015fully}
J.~Long, E.~Shelhamer, T.~Darrell, Fully convolutional networks for semantic
  segmentation, in: CVPR, 2015, pp. 3431--3440.

\bibitem{shelhamer2017fully}
E.~Shelhamer, J.~Long, T.~Darrell, Fully convolutional networks for semantic
  segmentation, {IEEE} Trans. Pattern Anal. Mach. Intell. 39~(4) (2017)
  640--651.
\newblock \href {https://doi.org/10.1109/TPAMI.2016.2572683}
  {\path{doi:10.1109/TPAMI.2016.2572683}}.

\bibitem{li2019deep}
Z.~Li, H.~Shen, Q.~Cheng, Y.~Liu, S.~You, Z.~He, Deep learning based cloud
  detection for medium and high resolution remote sensing images of different
  sensors, ISPRS Journal of Photogrammetry and Remote Sensing 150 (2019)
  197--212.
\newblock \href {https://doi.org/10.1016/j.isprsjprs.2019.02.017}
  {\path{doi:10.1016/j.isprsjprs.2019.02.017}}.

\bibitem{badrinarayanan2017segnet}
V.~Badrinarayanan, A.~Kendall, R.~Cipolla, {SegNet}: A deep convolutional
  encoder-decoder architecture for image segmentation, {IEEE} Trans. Pattern
  Anal. Mach. Intell. 39~(12) (2017) 2481--2495.

\bibitem{mohajerani2018cloud}
S.~Mohajerani, T.~A. Krammer, P.~Saeedi, A cloud detection algorithm for remote
  sensing images using fully convolutional neural networks, in: 2018 IEEE 20th
  International Workshop on Multimedia Signal Processing (MMSP), 2018.
\newblock \href {https://doi.org/10.1109/MMSP.2018.8547095}
  {\path{doi:10.1109/MMSP.2018.8547095}}.

\bibitem{yang2019cdnet}
J.~Yang, J.~Guo, H.~Yue, Z.~Liu, H.~Hu, K.~Li, {CDnet: CNN-Based Cloud
  Detection for Remote Sensing Imagery}, {IEEE} Trans. Geosci. Remote Sens.
  57~(8) (2019) 6195--6211.
\newblock \href {https://doi.org/10.1109/TGRS.2019.2904868}
  {\path{doi:10.1109/TGRS.2019.2904868}}.

\bibitem{zhang2021cnn}
J.~Zhang, Y.~Wang, H.~Wang, J.~Wu, Y.~Li, {CNN} cloud detection algorithm based
  on channel and spatial attention and probabilistic upsampling for remote
  sensing image, {IEEE} Trans. Geosci. Remote Sens. (2021) 1--13Early access.
\newblock \href {https://doi.org/10.1109/TGRS.2021.3105424}
  {\path{doi:10.1109/TGRS.2021.3105424}}.

\bibitem{ronneberger2015u-net}
O.~Ronneberger, P.~Fischer, T.~Brox, {U-Net}: Convolutional networks for
  biomedical image segmentation, in: MICCAI, 2015.

\bibitem{griffin2003cloud}
M.~Griffin, H.~Burke, D.~Mandl, J.~Miller, {Cloud cover detection algorithm for
  EO-1 Hyperion imagery}, in: IEEE International Geoscience and Remote Sensing
  Symposium, Vol.~1, 2003, pp. 86--89.
\newblock \href {https://doi.org/10.1109/IGARSS.2003.1293687}
  {\path{doi:10.1109/IGARSS.2003.1293687}}.

\bibitem{du2022adversarial}
A.~Du, Y.~W. Law, M.~Sasdelli, B.~Chen, K.~Clarke, M.~Brown, T.-J. Chin,
  Adversarial attacks against a satellite-borne multispectral cloud detector,
  in: 2022 International Conference on Digital Image Computing: Techniques and
  Applications (DICTA), 2022, arXiv version at
  \url{https://arxiv.org/abs/2112.01723}.
\newblock \href {https://doi.org/10.1109/DICTA56598.2022.10034592}
  {\path{doi:10.1109/DICTA56598.2022.10034592}}.

\bibitem{ruzicka2023fast}
V.~Růžička, G.~Mateo-García, C.~Bridges, C.~Brunskill, C.~Purcell,
  N.~Longépé, A.~Markham, Fast model inference and training on-board of
  satellites, in: International Geoscience and Remote Sensing Symposium, 2023,
  preprint arXiv:2307.08700.

\bibitem{ruzicka2022ravaen}
V.~Růžička, A.~Vaughan, D.~De~Martini, J.~Fulton, V.~Salvatelli, C.~Bridges,
  G.~Mateo-Garcia, V.~Zantedeschi, {RaVÆn}: unsupervised change detection of
  extreme events using {ML} on-board satellites, Scientific Reports 12~(1)
  (2022) 16939.
\newblock \href {https://doi.org/10.1038/s41598-022-19437-5}
  {\path{doi:10.1038/s41598-022-19437-5}}.

\bibitem{dorbit2023dashing}
{D-Orbit}, \href{https://www.dorbit.space/media/3/97.pdf}{{Dashing through the
  Stars Mission Booklet}} (2023).
\newline\urlprefix\url{https://www.dorbit.space/media/3/97.pdf}

\bibitem{kingma2022autoencoding}
D.~P. Kingma, M.~Welling, Auto-encoding variational bayes, arXiv preprint
  arXiv:1312.6114, version 11 (2022).

\bibitem{mateogarcia2023inorbit}
G.~Mateo-Garcia, J.~Veitch-Michaelis, C.~Purcell, N.~Longepe, S.~Reid,
  A.~Anlind, F.~Bruhn, J.~Parr, P.~P. Mathieu, In-orbit demonstration of a
  re-trainable machine learning payload for processing optical imagery,
  Scientific Reports 13~(1) (2023) 10391.
\newblock \href {https://doi.org/10.1038/s41598-023-34436-w}
  {\path{doi:10.1038/s41598-023-34436-w}}.

\bibitem{zou2022learning}
Y.~Zou, Z.~Zhang, C.-L. Li, H.~Zhang, T.~Pfister, J.-B. Huang, Learning
  instance-specific adaptation for cross-domain segmentation, in: S.~Avidan,
  G.~Brostow, M.~Ciss{\'e}, G.~M. Farinella, T.~Hassner (Eds.), Computer Vision
  -- ECCV 2022, Springer Nature Switzerland, Cham, 2022, pp. 459--476.

\bibitem{liang2023comprehensive}
J.~Liang, R.~He, T.~Tan, A comprehensive survey on test-time adaptation under
  distribution shifts, arXiv preprint arXiv:2303.15361 (2023).

\bibitem{farahani2021brief}
A.~Farahani, S.~Voghoei, K.~Rasheed, H.~R. Arabnia, A brief review of domain
  adaptation, in: R.~Stahlbock, G.~M. Weiss, M.~Abou-Nasr, C.-Y. Yang, H.~R.
  Arabnia, L.~Deligiannidis (Eds.), Advances in Data Science and Information
  Engineering, Springer International Publishing, Cham, 2021, pp. 877--894.

\bibitem{liu2022deep}
X.~Liu, C.~Yoo, F.~Xing, H.~Oh, G.~E. Fakhri, J.-W. Kang, J.~Woo, Deep
  unsupervised domain adaptation: A review of recent advances and perspectives,
  APSIPA Transactions on Signal and Information Processing 11~(1) (2022).
\newblock \href {https://doi.org/10.1561/116.00000192}
  {\path{doi:10.1561/116.00000192}}.

\bibitem{peng2022domain}
J.~Peng, Y.~Huang, W.~Sun, N.~Chen, Y.~Ning, Q.~Du, Domain adaptation in remote
  sensing image classification: A survey, {IEEE} J. Sel. Topics Appl. Earth
  Observ. Remote Sens. 15 (2022) 9842--9859.
\newblock \href {https://doi.org/10.1109/JSTARS.2022.3220875}
  {\path{doi:10.1109/JSTARS.2022.3220875}}.

\bibitem{zhang2022transfer}
L.~Zhang, X.~Gao, Transfer adaptation learning: A decade survey, {IEEE} Trans.
  Neural Netw. Learn. Syst. (2022) 1--22\href
  {https://doi.org/10.1109/TNNLS.2022.3183326}
  {\path{doi:10.1109/TNNLS.2022.3183326}}.

\bibitem{fang2023source}
Y.~Fang, P.-T. Yap, W.~Lin, H.~Zhu, M.~Liu, Source-free unsupervised domain
  adaptation: A survey, arXiv preprint arXiv:2301.00265 (2023).

\bibitem{singhal2023domain}
P.~Singhal, R.~Walambe, S.~Ramanna, K.~Kotecha, Domain adaptation: Challenges,
  methods, datasets, and applications, {IEEE} Access 11 (2023) 6973--7020.
\newblock \href {https://doi.org/10.1109/ACCESS.2023.3237025}
  {\path{doi:10.1109/ACCESS.2023.3237025}}.

\bibitem{yu2023comprehensive}
Z.~Yu, J.~Li, Z.~Du, L.~Zhu, H.~T. Shen, A comprehensive survey on source-free
  domain adaptation, arXiv preprint arXiv:2302.11803 (2023).

\bibitem{kellenberger2021deep}
B.~Kellenberger, O.~Tasar, B.~Bhushan~Damodaran, N.~Courty, D.~Tuia, Deep
  Domain Adaptation in Earth Observation, John Wiley \& Sons, Ltd, 2021, Ch.~7,
  pp. 90--104.
\newblock \href {https://doi.org/10.1002/9781119646181.ch7}
  {\path{doi:10.1002/9781119646181.ch7}}.

\bibitem{zhou2023domain}
K.~Zhou, Z.~Liu, Y.~Qiao, T.~Xiang, C.~C. Loy, Domain generalization: A survey,
  {IEEE} Trans. Pattern Anal. Mach. Intell. 45~(4) (2023) 4396--4415.
\newblock \href {https://doi.org/10.1109/TPAMI.2022.3195549}
  {\path{doi:10.1109/TPAMI.2022.3195549}}.

\bibitem{lucas2023bayesian}
B.~Lucas, C.~Pelletier, D.~Schmidt, G.~I. Webb, F.~Petitjean, A
  {Bayesian}-inspired, deep learning-based, semi-supervised domain adaptation
  technique for land cover mapping, Machine Learning 112~(6) (2023) 1941--1973.
\newblock \href {https://doi.org/10.1007/s10994-020-05942-z}
  {\path{doi:10.1007/s10994-020-05942-z}}.

\bibitem{shendryk2019deep}
Y.~Shendryk, Y.~Rist, C.~Ticehurst, P.~Thorburn, {Deep learning for multi-modal
  classification of cloud, shadow and land cover scenes in PlanetScope and
  Sentinel-2 imagery}, ISPRS Journal of Photogrammetry and Remote Sensing 157
  (2019) 124--136.
\newblock \href {https://doi.org/10.1016/j.isprsjprs.2019.08.018}
  {\path{doi:10.1016/j.isprsjprs.2019.08.018}}.

\bibitem{mateogarcia2020transferring}
G.~Mateo-Garc{\'\i}a, V.~Laparra, D.~L{\'o}pez-Puigdollers, L.~G{\'o}mez-Chova,
  {Transferring deep learning models for cloud detection between Landsat-8 and
  Proba-V}, ISPRS Journal of Photogrammetry and Remote Sensing 160 (2020)
  1--17.
\newblock \href {https://doi.org/10.1016/j.isprsjprs.2019.11.024}
  {\path{doi:10.1016/j.isprsjprs.2019.11.024}}.

\bibitem{segalrozenhaimer2020cloud}
M.~Segal-Rozenhaimer, A.~Li, K.~Das, V.~Chirayath, Cloud detection algorithm
  for multi-modal satellite imagery using convolutional neural-networks
  ({CNN}), Remote Sensing of Environment 237 (2020) 111446.
\newblock \href {https://doi.org/10.1016/j.rse.2019.111446}
  {\path{doi:10.1016/j.rse.2019.111446}}.

\bibitem{chen2018deeplab}
L.-C. Chen, G.~Papandreou, I.~Kokkinos, K.~Murphy, A.~L. Yuille, {DeepLab:
  Semantic Image Segmentation with Deep Convolutional Nets, Atrous Convolution,
  and Fully Connected CRFs}, {IEEE} Trans. Pattern Anal. Mach. Intell. 40~(4)
  (2018) 834--848.
\newblock \href {https://doi.org/10.1109/TPAMI.2017.2699184}
  {\path{doi:10.1109/TPAMI.2017.2699184}}.

\bibitem{mateogarcia2021cross}
G.~Mateo-Garc{\'\i}a, V.~Laparra, D.~L{\'o}pez-Puigdollers, L.~G{\'o}mez-Chova,
  Cross-sensor adversarial domain adaptation of landsat-8 and proba-v images
  for cloud detection, {IEEE} J. Sel. Topics Appl. Earth Observ. Remote Sens.
  14 (2021) 747--761.
\newblock \href {https://doi.org/10.1109/JSTARS.2020.3031741}
  {\path{doi:10.1109/JSTARS.2020.3031741}}.

\bibitem{zhu2017unpaired}
J.-Y. Zhu, T.~Park, P.~Isola, A.~A. Efros, Unpaired image-to-image translation
  using cycle-consistent adversarial networks, in: Proceedings of the IEEE
  International Conference on Computer Vision (ICCV), 2017.
\newblock \href {https://doi.org/10.1109/ICCV.2017.244}
  {\path{doi:10.1109/ICCV.2017.244}}.

\bibitem{xu2022source}
Z.~Xu, W.~Wei, L.~Zhang, J.~Nie, Source-free domain adaptation for cross-scene
  hyperspectral image classification, in: IGARSS 2022 - 2022 IEEE International
  Geoscience and Remote Sensing Symposium, 2022, pp. 3576--3579.
\newblock \href {https://doi.org/10.1109/IGARSS46834.2022.9883053}
  {\path{doi:10.1109/IGARSS46834.2022.9883053}}.

\bibitem{li2017spectral}
Y.~Li, H.~Zhang, Q.~Shen, Spectral–spatial classification of hyperspectral
  imagery with {3D} convolutional neural network, Remote Sensing 9~(1) (2017).
\newblock \href {https://doi.org/10.3390/rs9010067}
  {\path{doi:10.3390/rs9010067}}.

\bibitem{qiu2021source}
Z.~Qiu, Y.~Zhang, H.~Lin, S.~Niu, Y.~Liu, Q.~Du, M.~Tan, Source-free domain
  adaptation via avatar prototype generation and adaptation, in: Proceedings of
  the Thirtieth International Joint Conference on Artificial Intelligence,
  {IJCAI-21}, 2021, pp. 2921--2927.
\newblock \href {https://doi.org/10.24963/ijcai.2021/402}
  {\path{doi:10.24963/ijcai.2021/402}}.

\bibitem{liu2021adversarial}
X.~Liu, Z.~Guo, S.~Li, F.~Xing, J.~You, C.-C.~J. Kuo, G.~El~Fakhri, J.~Woo,
  Adversarial unsupervised domain adaptation with conditional and label shift:
  Infer, align and iterate, in: Proceedings of the IEEE/CVF International
  Conference on Computer Vision (ICCV), 2021, pp. 10367--10376.
\newblock \href {https://doi.org/10.1109/ICCV48922.2021.01020}
  {\path{doi:10.1109/ICCV48922.2021.01020}}.

\bibitem{kullback1951information}
S.~Kullback, R.~A. Leibler, On information and sufficiency, The Annals of
  Mathematical Statistics 22~(1) (1951) 79--86.

\bibitem{wang2021tent}
D.~Wang, E.~Shelhamer, S.~Liu, B.~Olshausen, T.~Darrell,
  \href{https://openreview.net/forum?id=uXl3bZLkr3c}{Tent: Fully test-time
  adaptation by entropy minimization}, in: ICLR, 2021.
\newline\urlprefix\url{https://openreview.net/forum?id=uXl3bZLkr3c}

\bibitem{ioffe2015batch}
S.~Ioffe, C.~Szegedy,
  \href{https://proceedings.mlr.press/v37/ioffe15.html}{Batch normalization:
  Accelerating deep network training by reducing internal covariate shift}, in:
  F.~Bach, D.~Blei (Eds.), Proceedings of the 32nd International Conference on
  Machine Learning, Vol.~37 of Proceedings of Machine Learning Research, PMLR,
  Lille, France, 2015, pp. 448--456.
\newline\urlprefix\url{https://proceedings.mlr.press/v37/ioffe15.html}

\bibitem{mirza2022norm}
M.~J. Mirza, J.~Micorek, H.~Possegger, H.~Bischof, The norm must go on: dynamic
  unsupervised domain adaptation by normalization, in: Proceedings of the
  IEEE/CVF Conference on Computer Vision and Pattern Recognition, 2022, pp.
  14765--14775.

\bibitem{furano2020towards}
G.~Furano, G.~Meoni, A.~Dunne, D.~Moloney, V.~Ferlet-Cavrois, A.~Tavoularis,
  J.~Byrne, L.~Buckley, M.~Psarakis, K.-O. Voss, et~al., Towards the use of
  artificial intelligence on the edge in space systems: Challenges and
  opportunities, IEEE Aerospace and Electronic Systems Magazine 35~(12) (2020)
  44--56.

\bibitem{papadimitriou2007tcp}
P.~Papadimitriou, V.~Tsaoussidis, On {TCP} performance over asymmetric
  satellite links with real-time constraints, Computer Communications 30~(7)
  (2007) 1451--1465.
\newblock \href {https://doi.org/10.1016/j.comcom.2006.12.030}
  {\path{doi:10.1016/j.comcom.2006.12.030}}.

\bibitem{2023pytorch}
{PyTorch Foundation}, \href{https://pytorch.org/}{{PyTorch}}, accessed 35 Apr
  2023 (2023).
\newline\urlprefix\url{https://pytorch.org/}

\bibitem{francis_alistair_2020_4172871}
A.~Francis, J.~Mrziglod, P.~Sidiropoulos, J.-P. Muller, Sentinel-2 cloud mask
  catalogue (version 1), Dataset under CC BY 4.0 license at
  {\small\url{https://doi.org/10.5281/zenodo.4172871}} (Nov. 2020).

\bibitem{2023usgs}
{USGS}, \href{https://earthexplorer.usgs.gov/}{{EarthExplorer}}, USGS: science
  for a changing world, accessed 3 Apr 2023 (2023).
\newline\urlprefix\url{https://earthexplorer.usgs.gov/}

\bibitem{he2016deep}
K.~He, X.~Zhang, S.~Ren, J.~Sun, Deep residual learning for image recognition,
  in: Proceedings of the IEEE conference on computer vision and pattern
  recognition, 2016, pp. 770--778.

\bibitem{sung2021training}
Y.-L. Sung, V.~Nair, C.~A. Raffel,
  \href{https://proceedings.neurips.cc/paper_files/paper/2021/file/cb2653f548f8709598e8b5156738cc51-Paper.pdf}{Training
  neural networks with fixed sparse masks} 34 (2021) 24193--24205.
\newline\urlprefix\url{https://proceedings.neurips.cc/paper_files/paper/2021/file/cb2653f548f8709598e8b5156738cc51-Paper.pdf}

\bibitem{2023onnxruntime}
{Microsoft}, \href{https://onnxruntime.ai/}{{ONNX Runtime}}, accessed 7 Jun
  2023 (2023).
\newline\urlprefix\url{https://onnxruntime.ai/}

\bibitem{li2017revisiting}
Y.~Li, N.~Wang, J.~Shi, J.~Liu, X.~Hou,
  \href{https://openreview.net/pdf?id=Hk6dkJQFx}{Revisiting batch normalization
  for practical domain adaptation}, in: ICLR workshop, 2017.
\newline\urlprefix\url{https://openreview.net/pdf?id=Hk6dkJQFx}

\bibitem{schneider2020improving}
S.~Schneider, E.~Rusak, L.~Eck, O.~Bringmann, W.~Brendel, M.~Bethge, Improving
  robustness against common corruptions by covariate shift adaptation, in:
  Advances in Neural Information Processing Systems, Vol.~33, Curran
  Associates, Inc., 2020, pp. 11539--11551.

\end{thebibliography}
